\pdfoutput=1
\documentclass[11pt]{article}

\usepackage[final]{acl}

\usepackage{times}
\usepackage{latexsym}
\usepackage[T1]{fontenc}
\usepackage[utf8]{inputenc}
\usepackage{microtype}
\usepackage{inconsolata}
\usepackage{graphicx}
\usepackage{booktabs}
\usepackage{enumitem}
\usepackage{amssymb}
\usepackage{multirow}
\usepackage{tikz}
\usetikzlibrary{arrows.meta,positioning,calc,shapes.geometric}
\usepackage{amsmath}
\usepackage{tabularx}
\usepackage{array}
\newcolumntype{L}[1]{>{\raggedright\arraybackslash}p{#1}}
\newcommand{\code}[1]{\texttt{\small #1}}
\usepackage{xcolor}
\usepackage{listings}
\usepackage[most]{tcolorbox}
\newtcolorbox{casebox}[1]{enhanced, colback=gray!4, colframe=gray!55, colbacktitle=gray!18, coltitle=black, boxrule=0.4pt, arc=1pt, left=4pt, right=4pt, fonttitle=\small\bfseries, fontupper=\footnotesize, title={#1}}
\newtcblisting{promptbox}[1]{enhanced, breakable, listing only, colback=gray!4, colframe=gray!55, colbacktitle=gray!18, coltitle=black, boxrule=0.4pt, arc=1pt, left=4pt, right=4pt, fonttitle=\small\bfseries, title={#1}, listing options={basicstyle=\ttfamily\scriptsize, breaklines=true, columns=fullflexible, keepspaces=true, upquote=true}}

\newcommand{\nstud}{100}
\newcommand{\nsess}{1{,}248}
\newcommand{\sessrange}{10--15}
\title{Psy-Chronicle: Synthesizing Campus Counseling Grounded in\\a Student's Semester Timeline}

\author{Chaogui Gou \\
  University of Science and Technology Beijing \\
  \texttt{ustbedwin@gmail.com}}

\begin{document}
\maketitle

\begin{abstract}
Psychological counseling unfolds over many sessions, and what a client brings to each session depends on what has happened in their life since the last one. Existing synthetic counseling corpora are mostly single-session, and multi-session ones advance along a therapy plan without modeling the client's life between sessions. We present Psy-Chronicle, an event-driven framework that grounds multi-session counseling synthesis in a student's semester timeline: each simulated student receives a profile and a semester-long temporal stress event graph, and counseling sessions are generated along this timeline with memory carried across them. Applied to campus counseling, it yields CPCD, an open Chinese dataset of 100 students and 1{,}248 sessions spanning a full semester. The dialogues follow their timelines: dependencies between events carry across sessions, and references to earlier sessions rarely introduce content absent from the trajectory. Under a published evaluation protocol, CPCD matches MusPsy, an existing multi-session corpus, in basic quality and exceeds all single-session corpora in working alliance. A model fine-tuned on CPCD connects each week's events to earlier ones more often than a model trained on the same data without the dependencies between events, while matching it when the client's situation does not change.
\end{abstract}

% ==================================================================
\section{Introduction}
% ==================================================================

Large language models (LLMs) are increasingly used for psychological support, and synthesized dialogues have become a main source of training data for this task~\citep{sun2021psyqa,liu2021esconv,qiu2024smile,zhang2024cpsycoun,lee2024cactus}.
Most of these resources treat counseling as a single conversation.
In practice, counseling unfolds over multiple sessions, in which the counselor tracks the client's progress and adjusts goals over time~\citep{beck2011cbt}.
Recent work has begun to synthesize multi-session counseling data~\citep{wang-etal-2026-psychological,zhou2025diacbt,liu2026dmtcbt,li2026longcounsel}.

In these corpora, however, progression across sessions is organized around the therapy course, and the client's life between sessions is at most described session by session rather than modeled as a timeline.
In reality, clients return to counseling because their lives go on: new events occur, earlier ones have lasting consequences, and the concerns brought to each session are shaped by what has happened since the last one.

Campus counseling is a typical example.
College students' stressors follow the academic calendar, from adjustment and midterms to finals and job hunting, and stress rises as the term progresses~\citep{wang2014studentlife}. A failed midterm may resurface weeks later as insomnia or a dormitory conflict.
Demand for support is high: about one in three first-year students screens positive for a common mental disorder~\citep{auerbach2018who}, and the prevalence of depression among Chinese university students is 28.4\%~\citep{gao2020prevalence}. Counseling resources, by contrast, are limited~\citep{moe2023action}.
Yet open data for this setting is scarce. Existing Chinese counseling corpora are general-purpose~\citep{qiu2024smile,zhang2024cpsycoun,qiu2025psydial}, and the only school counseling corpus we are aware of covers single conversations~\citep{zhao2026tudial}.

In this paper, we propose \textbf{Psy-Chronicle}, an event-driven framework for synthesizing multi-session counseling dialogues, and apply it to campus counseling.
For each simulated student, Psy-Chronicle builds a profile and a semester-long \emph{temporal stress event graph}, then generates CBT-guided counseling sessions along this timeline with memory carried across sessions.
Based on Psy-Chronicle, we construct the CPCD dataset and train a counseling model, CPCD-Chat.
Our contributions are as follows:
\begin{itemize}[leftmargin=*,itemsep=1pt,topsep=2pt]
  \item We propose Psy-Chronicle, an event-driven framework that grounds multi-session counseling synthesis in a student's semester timeline.
  \item We construct and release CPCD, to our knowledge the first open Chinese multi-session dataset for campus counseling, with \nstud{} students and \nsess{} sessions spanning a full semester.
  \item We develop CPCD-Chat, a counseling model fine-tuned on CPCD, and introduce a dynamic evaluation protocol in which the client's life changes between sessions. Under this protocol, the dependencies between events in the training data improve how the model connects sessions to earlier events, with no loss under the static protocol of prior work.
\end{itemize}

% ==================================================================
\section{Related Work}
\label{sec:related}
% ==================================================================

\begin{table}[t]
  \centering
  \small
  \setlength{\tabcolsep}{4pt}
  \begin{tabular}{lcccc}
    \toprule
    \textbf{Dataset} & \textbf{Lang.} & \textbf{Setting} & \textbf{Sess.} & \textbf{Timeline} \\
    \midrule
    PsyQA          & Zh & General & QA & $\times$ \\
    ESConv        & En & General & 1  & $\times$ \\
    SMILE          & Zh & General & 1  & $\times$ \\
    CPsyCoun  & Zh & General & 1  & $\times$ \\
    CACTUS        & En & General & 1  & $\times$ \\
    PsyDial      & Zh & General & 1  & $\times$ \\
    MusPsy       & En & General & 6.2 & $\times$ \\
    DiaCBT       & En & General & 5  & $\times$ \\
    LongCounsel-8 & En & General & 5 & $\times$ \\
    \midrule
    \textbf{CPCD (ours)}                      & Zh & Campus  & \sessrange & \checkmark \\
    \bottomrule
  \end{tabular}
  \caption{Comparison of counseling dialogue datasets. \textbf{Sess.}: sessions per client. \textbf{Timeline}: life events between sessions are organized by date and linked by dependencies.}
  \label{tab:compare}
\end{table}

\paragraph{Psychological counseling datasets.}
Early resources collected single-turn question answering~\citep{sun2021psyqa} and emotional support conversations~\citep{liu2021esconv}.
With LLMs, multi-turn counseling dialogues have been synthesized by expanding question-answer pairs~\citep{qiu2024smile}, reconstructing counseling reports~\citep{zhang2024cpsycoun}, applying CBT techniques~\citep{lee2024cactus}, or refining generated dialogues toward real-world data~\citep{shi2025cpsdd}, while PsyDial~\citep{qiu2025psydial} provides long single sessions reconstructed from real counseling.
These corpora support counseling models such as ChatCounselor~\citep{liu2023chatcounselor} and PsyLLM~\citep{hu2025psyllm}.
They mostly target general counseling within a single session. TuDialCorpus~\citep{zhao2026tudial} addresses school counseling, but likewise models one conversation.

\paragraph{Multi-session counseling synthesis.}
Recent work generates counseling that spans multiple sessions.
MusPsy~\citep{wang-etal-2026-psychological} extracts per-session goals and client states from published case reports.
DiaCBT~\citep{zhou2025diacbt} guides five-session CBT courses with cognitive conceptualization diagrams, DMT-CBT~\citep{liu2026dmtcbt} tracks structured therapeutic states across sessions, and LongCounsel-8~\citep{li2026longcounsel} simulates five-session trajectories from prescribed depression severity.
Graph2Counsel~\citep{mandal2026graph2counsel} grounds counseling in a graph of the client's thoughts, emotions, and behaviors.
In these works, progression is organized around the therapy course or the client's clinical state. Life events, when present, are described per session and are not linked into a timeline with dates and dependencies.
Table~\ref{tab:compare} summarizes these differences.

\paragraph{Long-term conversation with event timelines.}
Open-domain dialogue research has studied conversations that span many sessions~\citep{xu2022msc,jang2023chronicles}.
LoCoMo~\citep{maharana2024locomo} grounds each speaker in a persona and a temporal event graph, and generates conversations of up to 32 sessions that follow the graph.
Counseling adds constraints that open-domain conversation lacks: the counselor must not see what the client has not disclosed, the pace of therapy must follow the client's distress, and references to earlier sessions must stay faithful to what the client said. In Psy-Chronicle, events become campus stressors with intensity and dependencies on earlier events, and they shape both what the student brings to each session and how counseling progresses.

% ==================================================================
\section{Psy-Chronicle}
\label{sec:method}
% ==================================================================
\begin{figure*}[t]
  \centering
  \resizebox{\textwidth}{!}{% Figure 1: Psy-Chronicle framework (TikZ). \input from acl_latex.tex
\begin{tikzpicture}[
  font=\sffamily\scriptsize, >={Stealth[length=4pt]},
  panel/.style={draw=black!35, rounded corners=3pt, fill=black!2, inner sep=4pt},
  ptitle/.style={font=\sffamily\footnotesize\bfseries, anchor=north west},
  field/.style={draw=black!25, fill=white, rounded corners=1.5pt, minimum width=3.0cm, minimum height=0.5cm, align=left, text width=2.85cm, inner sep=2pt},
  ev/.style={circle, draw=white, line width=0.6pt, minimum size=0.42cm, inner sep=0pt, font=\sffamily\tiny\bfseries, text=white},
  agent/.style={draw=black!45, fill=white, rounded corners=2pt, minimum width=1.55cm, minimum height=0.62cm, align=center},
  sees/.style={draw=black!20, fill=white, rounded corners=1.5pt, align=left, inner sep=2.5pt, text=black!80},
  dep/.style={->, draw=black!60, line width=0.55pt},
  flow/.style={->, draw=black!70, line width=0.9pt}]
\definecolor{cAca}{HTML}{3F6D96}\definecolor{cInt}{HTML}{6E5A96}\definecolor{cCar}{HTML}{217F6E}
\definecolor{cFam}{HTML}{A9662C}\definecolor{cHea}{HTML}{AC3F52}

% ---------------- Panel A: profile
\node[panel, minimum width=3.4cm, minimum height=4.0cm, anchor=south west] (PA) at (0,0) {};
\node[ptitle] at ($(PA.north west)+(0.05,-0.05)$) {(1) Student profile};
\node[field] (f1) at ($(PA.north)+(0,-0.85)$) {\textbf{Demographics}\\[-1pt] gender, year, major};
\node[field, below=0.2cm of f1] (f2) {\textbf{Personality}\\[-1pt] type, traits, coping style};
\node[field, below=0.2cm of f2] (f3) {\textbf{Background}\\[-1pt] family, social support};
\node[field, below=0.2cm of f3] (f4) {\textbf{Core conflict}\\[-1pt] underlying inner tension};

% ---------------- Panel B: event graph
\node[panel, minimum width=6.6cm, minimum height=4.0cm, anchor=south west] (PB) at (3.8,0) {};
\node[ptitle] at ($(PB.north west)+(0.05,-0.05)$) {(2) Temporal stress event graph};
\def\xo{4.15}\def\wk{0.37}\def\yb{0.55}
\draw[->, black!50] (\xo,\yb) -- (\xo+16*\wk+0.15,\yb) node[right, black!60, font=\sffamily\tiny]{wk};
\foreach \w in {1,4,8,12,16}{\draw[black!50] (\xo+\w*\wk-0.5*\wk,\yb) -- ++(0,-0.06) node[below, font=\sffamily\tiny, black!60]{\w};}
% events: name/week/stress/colour
\foreach \n/\w/\s/\c in {E1/2/5/cAca, E2/4/4/cInt, E3/7/8/cAca, E4/10/9/cHea, E5/12/6/cFam, E6/15/7/cCar}{
  \fill[\c!35] (\xo+\w*\wk-0.5*\wk-0.09,\yb) rectangle ++(0.18,\s*0.09);
  \node[ev, fill=\c] (\n) at (\xo+\w*\wk-0.5*\wk, \yb+\s*0.09+0.42) {\n};}
\draw[dep] (E1) to[bend left=38] (E3);
\draw[dep] (E2) to[bend left=40] (E4);
\draw[dep] (E3) to[bend left=35] (E4);
\draw[dep] (E4) to[bend left=30] (E6);
\draw[dep] (E5) to[bend left=35] (E6);
\node[font=\sffamily\tiny, black!65, anchor=west] at (\xo, 3.35) {arcs: \emph{depends on}\quad bars: stress level (1--10)};
\node[font=\sffamily\tiny, anchor=west] at (\xo, 3.08) {%
  \textcolor{cAca}{$\bullet$}\,academic\ \textcolor{cInt}{$\bullet$}\,interpersonal\ \textcolor{cCar}{$\bullet$}\,career\ \textcolor{cFam}{$\bullet$}\,family\ \textcolor{cHea}{$\bullet$}\,health};

% ---------------- Panel C: sessions
\node[panel, minimum width=6.3cm, minimum height=4.0cm, anchor=south west] (PC) at (10.8,0) {};
\node[ptitle] at ($(PC.north west)+(0.05,-0.05)$) {(3) Multi-session counseling};
\node[agent] (stu) at (12.45,2.75) {Student\\agent};
\node[agent] (cou) at (15.45,2.75) {Counselor\\agent};
\draw[<->, black!60, line width=0.7pt] (stu) -- node[above, align=center]{session $t$} (cou);
\node[sees, below=0.12cm of stu] (sS) {full profile\\event $t$ + antecedents};
\node[sees, below=0.12cm of cou] (sC) {no core conflict\\session goal, CBT stage};
\node[draw=black!45, fill=white, cylinder, shape border rotate=90, aspect=0.25, minimum width=1.9cm, minimum height=0.75cm, align=center] (mem) at (13.95,0.6) {memory\\(summaries)};
\draw[->, black!60] (cou.east) -- (16.95,2.75) |- node[pos=0.75, above]{write} (mem.east);
\draw[->, black!60] (mem.west) -| node[pos=0.25, above]{read at $t{+}1$} (11.0,2.75) -- (stu.west);

% ---------------- flows between panels
\draw[flow] (PA.east) -- (PB.west);
\draw[flow] (PB.east) -- (PC.west);
\end{tikzpicture}}
  \caption{Overview of Psy-Chronicle. Each simulated student is given a profile and a semester-long temporal stress event graph. Counseling sessions are generated in chronological order, one per event, and a structured memory is carried from each session to the next.}
  \label{fig:framework}
\end{figure*}

Psy-Chronicle synthesizes a semester of counseling for each simulated student in three steps (Figure~\ref{fig:framework}). Construction details are given in Appendices~\ref{app:profile} and~\ref{app:event}.

\paragraph{Student profiles.}
Each profile describes a student's demographics, personality, family and social background, and a core psychological conflict tied to a primary stress domain.
Categorical attributes are assigned programmatically to cover a full factorial of gender, academic year, and five stress domains, and an LLM writes the remaining text. All generation in Psy-Chronicle uses DeepSeek-V4-Flash.

\paragraph{Temporal stress event graph.}
From the profile, the LLM generates 10--15 dated campus events spanning a 16-week semester. Each event has a domain, a stress level, and a psychological impact, and is linked to the earlier events it depends on.
The structure of each graph, including its time span, domain quota, and one of five stress trajectory archetypes, is sampled before generation, and every graph is validated and revised automatically.

\paragraph{Multi-session counseling.}
A student agent and a counselor agent hold one session per event in chronological order.
The student agent sees the full profile, the current event and the events it depends on, while the counselor agent sees the demographics, the memory, and a session goal derived from the event graph, but never the core conflict, background, or personality, which must emerge through conversation.
Each session follows the standard CBT session structure~\citep{beck2011cbt}, and across sessions the counselor progresses through the stages of a CBT course using named CBT techniques from the catalogue of \citet{lee2024cactus}, slowing down at high-stress events and advancing at low-stress ones.
After each session, a structured summary is written to a memory that both agents read in later sessions. All generation prompts are given in Appendix~\ref{app:prompts}.

% ==================================================================
\section{The CPCD Dataset}
\label{sec:dataset}
% ==================================================================

\begin{table}[t]
  \centering
  \small
  \begin{tabular}{lr}
    \toprule
    Students / sessions            & \nstud{} / \nsess{} \\
    Utterances                     & 54{,}176 \\
    Characters                     & 4{,}252{,}810 \\
    Sessions per student           & 12.5 (\sessrange) \\
    Turns per session              & 21.2 (16--27) \\
    Characters per session         & 3{,}408 \\
    \midrule
    Events per graph               & 12.5 (10--15) \\
    Events with an upstream event  & 90.1\% \\
    Longest dependency chain       & 9 (4--12) \\
    \bottomrule
  \end{tabular}
  \caption{Statistics of CPCD. Values in parentheses are ranges; chain length is the median.}
  \label{tab:stats}
\end{table}

\begin{figure*}[t]
  \centering
  \includegraphics[width=\textwidth]{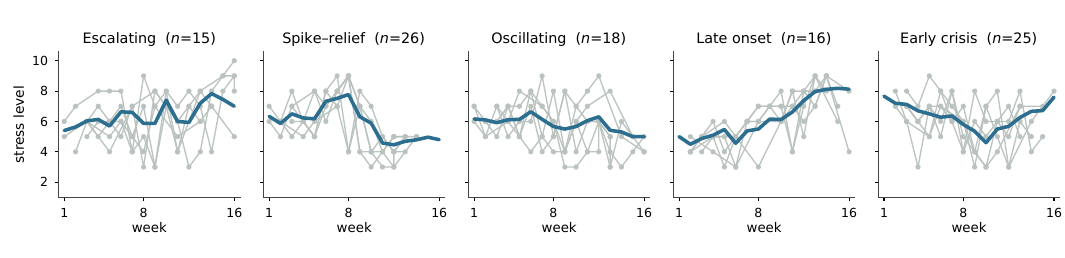}
  \caption{Stress trajectories of the 100 event graphs, grouped by the sampled archetype. Grey lines are individual students; the blue line is the archetype mean.}
  \label{fig:shapes}
\end{figure*}

CPCD contains \nstud{} simulated students, each with a full semester of counseling (Table~\ref{tab:stats}).

\paragraph{Profiles and event graphs.}
The profiles fill all 40 cells of the gender $\times$ year $\times$ domain design with two to three students each, and cover all 28 majors and 16 personality types.
The five trajectory archetypes are realized as distinct semester-long dynamics rather than as labels (Figure~\ref{fig:shapes}).
Events form connected chains rather than isolated incidents: 90.1\% of events depend on an earlier one, and the longest dependency chain per student has a median length of 9.
The timelines also follow the academic calendar, although no exam schedule is given in the prompt: events mentioning midterms peak in week 8, and 76\% of those mentioning final exams fall in weeks 13--16.
Appendix~\ref{app:case} follows one student through the semester and shows how the event graph, the profile, and the memory surface in the dialogue.

% ==================================================================
\section{Experiments}
\label{sec:exp}
% ==================================================================

Our experiments address five questions. The first three concern the data and the last two concern models trained on it.
\textbf{RQ1}: Do the synthesized dialogues realize what the profiles and event graphs specify?
\textbf{RQ2}: Is CPCD at least on par with existing counseling corpora in quality?
\textbf{RQ3}: Which properties of the dialogues does the timeline account for, and which the profile?
\textbf{RQ4}: Are models trained on CPCD better at counseling clients whose lives change between sessions?
\textbf{RQ5}: Does this ability come at the cost of general counseling quality?

\subsection{Setup}
\label{sec:setup}

\paragraph{Quality metrics.}
An LLM judge rates whole counseling sessions on two groups of metrics, following the protocol that \citet{wang-etal-2026-psychological} introduced to compare counseling corpora; the same metrics are later applied to sessions produced by trained models.

\paragraph{Basic metrics.}
Four dimensions assess the counseling conversation: \emph{helpfulness}, the applicability of the counselor's explanations and suggestions; \emph{empathy}, the counselor's ability to understand and share the client's feelings; \emph{guidance}, the availability and specificity of practical advice; and \emph{coherence}, the logical consistency of the conversation.

\paragraph{Working alliance.}
The working alliance comprises agreement between client and counselor on the goals of counseling, agreement on the tasks used to reach them, and the emotional bond between them~\citep{horvath1989wai}. It is among the most consistent predictors of counseling outcome~\citep{horvath1991alliance,horvath2011alliance} and applies across forms of counseling, which makes it a common yardstick for different corpora. We use the 12-item short revised Working Alliance Inventory (WAI-SR)~\citep{hatcher2006waisr,munder2010waisr}, which the judge completes from the client's perspective, and report its goal, task, and bond subscales and their average.

\paragraph{Judging procedure.}
For each corpus, we sample 100 clients and one session per client. The full transcript of the session is given to GPT-4o at temperature 0, which returns an integer score from 1 to 5 for each item without explanation. Prompts follow the language of the corpus, and Chinese corpora are rated with the Chinese version of the WAI-SR. All agents in Psy-Chronicle use DeepSeek-V4-Flash, so the judge comes from a different model family than the generator.
We score every corpus with our own implementation rather than reusing published scores. On MusPsy, it reproduces the published basic scores within 0.01 and the WAI average within 0.05, while the WAI subscales differ between pipelines (Appendix~\ref{app:calib}).

\paragraph{Models.}
We fine-tune Qwen2.5-7B-Instruct~\citep{qwen2025qwen25} on different training data while keeping everything else fixed: \emph{CPCD-Chat} on CPCD, \emph{w/o dependencies}, \emph{w/o timeline}, and \emph{w/o profile} on the three ablated versions of CPCD (Section~\ref{sec:rq3}), and two external baselines on the Chinese counseling corpora SMILE and CPsyCoun, down-sampled to the same number of counselor tokens as CPCD, and a multi-session baseline, \emph{MusPsy}, trained on the MusPsy corpus, whose sessions carry memory and plans but no timeline. We also report the untuned base model (\emph{Base}). All models thus share the same base model and differ only in their training data.

\paragraph{Training.}
A counselor who follows a student over a semester keeps notes after each session, plans the next one, and talks guided by both. We train these roles as three tasks, which keeps inputs short across sessions and lets the evaluation fix memory and plans while comparing responses: updating the memory from the previous memory and a session transcript, planning the next session from the memory, and generating the next counselor utterance from the memory, the plan, and the dialogue so far. Inputs and targets contain only what the counselor could know at that point: as a CPCD session centers on an event that has not yet happened when the previous one ends, plans cover only the course of therapy, and the week's event enters through the client (Appendix~\ref{app:training}). Single-session corpora yield only the last task, and MusPsy is translated into Chinese before training. All models are trained with LoRA~\citep{hu2022lora} (rank 32, $\alpha=16$, learning rate $2\times10^{-4}$, 2 epochs) using LlamaFactory~\citep{zheng2024llamafactory}; CPCD-Chat, w/o dependencies, and w/o timeline are trained with three random seeds (Appendix~\ref{app:training}).

\paragraph{Statistics.}
Model comparisons are paired by client, week, and simulation seed. Confidence intervals are 95\% bootstrap intervals with resampling by client (1{,}000 resamples). Equivalence is tested with two one-sided tests (TOST)~\citep{schuirmann1987tost} with a margin of $\pm0.15$ on the 1--5 scale, fixed in advance at three times the reproduction error of our scoring pipeline on the WAI average; as basic scores approach the ceiling, equivalence claims rest mainly on the WAI. Scores of multi-seed models are averaged over seeds before bootstrapping, and seed variation is reported separately. Secondary metrics are corrected with the Holm procedure.

\subsection{RQ1: Do the Dialogues Follow Their Timelines?}
\label{sec:rq1}

\begin{figure}[t]
  \centering
  \includegraphics[width=\columnwidth]{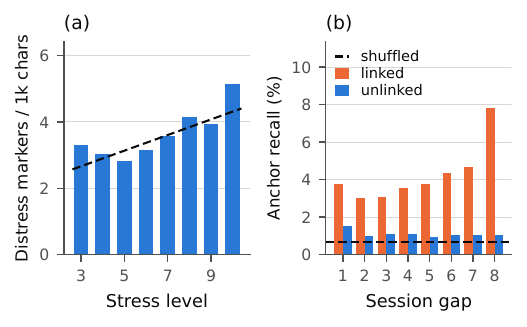}
  \caption{Dialogues follow the timeline. (a) Distress markers in student utterances by the stress level of the session's event; the dashed line is a linear fit. (b) Recall of an earlier event's content in a later session, for session pairs linked or unlinked by a dependency edge, by session gap; the dashed line is a cross-student shuffled baseline.}
  \label{fig:conformance}
\end{figure}

Because profiles and event graphs are fixed before any dialogue is generated, they serve as a reference against which the dialogues can be checked without human annotation.
Where a property could arise by chance, we compare it with a permutation null of 1{,}000 resamples (Figure~\ref{fig:conformance}; resampling units and full results in Appendix~\ref{app:conformance}). Checks of mechanisms that the pipeline sets directly, such as session length, serve only as sanity checks and are reported in the appendix.

\emph{Dependency propagation.}
For each event, we take as anchor terms the character 3- to 4-grams of its description that occur in no other event of the same student, and measure the share of them that reappear in a later session.
Session pairs linked by a dependency edge recall 3.53\% of the earlier event's anchors, against 1.10\% for unlinked pairs at the same session gap and 0.69\% for pairs drawn from different students ($p=0.001$, permutation stratified by gap). The difference holds at all eight gaps.

\emph{Grounded recall.}
91.8\% of non-first sessions refer back to an earlier session. To check that these references do not invent content, we trace the counselor's quotations attributed to the student in a random sample of 25 students. Of 3{,}674 quotations, 92.4\% match the student's own words. Of the 109 presented as coming from an earlier session, 45 match earlier wording and 18 match nothing; none of these 18 introduces a past episode absent from the trajectory (Appendix~\ref{app:conformance}).

\emph{Information asymmetry.}
The counselor prompt has no field for the core conflict, background, or personality (Appendix~\ref{app:prompt-counselor}); as a check on the generated text, no counselor turn contains a substring of the core conflict.

\emph{Stress conditioning.}
The student agent is instructed to express distress in line with the event's stress level; the density of distress and absolutist expressions in its speech (e.g., ``collapse'', ``always'') correlates weakly with it ($r=0.153$, $p<0.001$).

\subsection{RQ2: Dataset Quality}
\label{sec:rq2}

\begin{table}[t]
  \centering
  \small
  \setlength{\tabcolsep}{3.5pt}
  \begin{tabular}{lcccccc}
    \toprule
    \textbf{Corpus} & \textbf{Hel.} & \textbf{Emp.} & \textbf{Gui.} & \textbf{Coh.} & \textbf{Avg.} & \textbf{WAI} \\
    \midrule
    SMILE      & 4.07 & 4.47 & 3.99 & 4.45 & 4.25 & 3.81 \\
    CPsyCoun   & 4.25 & 4.45 & 4.08 & 4.86 & 4.41 & 3.84 \\
    CACTUS     & 4.60 & 4.88 & 4.55 & 4.99 & 4.76 & 3.95 \\
    MusPsy     & 4.98 & 4.99 & 4.98 & 5.00 & 4.99 & 4.36 \\
    \midrule
    \textbf{CPCD} & 4.93 & 4.98 & 4.90 & 4.99 & 4.95 & 4.22 \\
    \bottomrule
  \end{tabular}
  \caption{Dataset quality, with every corpus scored by our implementation of the protocol (GPT-4o judge, 100 sessions per corpus). Hel.: helpfulness; Emp.: empathy; Gui.: guidance; Coh.: coherence; WAI: average over the three subscales.}
  \label{tab:quality}
\end{table}

Table~\ref{tab:quality} compares CPCD with four public counseling corpora scored by the same pipeline: the single-session SMILE~\citep{qiu2024smile}, CPsyCoun~\citep{zhang2024cpsycoun}, and CACTUS~\citep{lee2024cactus}, and the multi-session MusPsy.
On the basic dimensions, CPCD scores 4.95 on average, on par with MusPsy (4.99) and above all single-session corpora (4.25--4.76).
On the working alliance, CPCD scores 4.22, higher than every single-session corpus (at most 3.95) and below MusPsy (4.36).
The gap lies in goal and task agreement (3.95 and 3.90 vs.\ 4.05 and 4.04), as MusPsy conditions each session on a goal from a case report, whereas CPCD derives goals from the event graph.
\emph{Human ratings.} Two psychology experts (bachelor's degree, at least two years of relevant experience), blind to the corpus, rated 100 CPCD and 100 MusPsy sessions on the same dimensions. They score lower than GPT-4o but keep its ordering (basic 4.74 vs.\ 4.80; WAI 4.15 vs.\ 4.31) and spread scores more widely, agreeing with the judge moderately at the session level (on CPCD, Spearman $\rho$ 0.48 basic, 0.57 WAI; Appendix~\ref{app:human}).
On the Cognitive Therapy Rating Scale (CTRS)~\citep{young1980ctrs}, an observer-rated measure of CBT competence (11 items, 0--6), CPCD averages 4.90 as rated by GPT-4o (Appendix~\ref{app:ctrs}).

\subsection{RQ3: What the Timeline and the Profile Contribute}
\label{sec:rq3}

To test whether each component contributes what it is designed to, we regenerate the 100 trajectories with the same agents and random seeds in three ablated settings.
The timeline has two parts, the order of events and the dependencies among them, which we remove in turn.
\emph{W/o dependencies} keeps every event in its original week but removes all dependency edges, so neither agent is told which earlier events the current one builds on.
\emph{W/o timeline} additionally reassigns the events to random weeks; since event descriptions were written in their original order, this also weakens narrative coherence, and we treat it as an upper bound.
\emph{W/o profile} keeps the event graph and only the demographics of the profile (gender, age, year, and major), removing the background, personality type and traits, coping style, and core conflict.
Each setting is compared with the full pipeline on the same students (Table~\ref{tab:ablation}; details in Appendix~\ref{app:ablation}). Dependency recall is measured against the original edges, excluding pairs whose antecedent no longer precedes the current event.

\begin{table}[t]
  \centering
  \small
  \footnotesize
  \setlength{\tabcolsep}{2pt}
  \begin{tabular}{lcccc}
    \toprule
    & \textbf{Full} & \textbf{w/o dep.} & \textbf{w/o time.} & \textbf{w/o prof.} \\
    \midrule
    Basic average        & 4.95 & 4.94 & 4.93   & 4.92 \\
    WAI goal             & 3.95 & 3.86$^{*}$ & 3.70$^{*}$ & 3.88 \\
    WAI task             & 3.90 & 3.83 & 3.65$^{*}$ & 3.84 \\
    WAI bond             & 4.80 & 4.80 & 4.78   & 4.60$^{*}$ \\
    \midrule
    Dep.\ recall (pp)   & +2.44$^{*}$ & +0.3 & +0.1 & +2.2$^{*}$ \\
    Cross-sess.\ ref.\ (\%) & 91.8 & 86 & 70 & 90 \\
    Cross-stud.\ sim.  & 0.011 & 0.011 & 0.011 & 0.015 \\
    \bottomrule
  \end{tabular}
  \caption{Ablation over the same 100 students. w/o dep.: w/o dependencies; w/o time.: w/o timeline; w/o prof.: w/o profile. Dep.\ recall: margin of linked over unlinked session pairs in anchor recall (Section~\ref{sec:rq1}); cross-session ref.: non-first sessions referring to an earlier one; cross-student sim.: mean character 4-gram Jaccard between student utterances of different students. $^{*}$: $p<0.01$ (paired Wilcoxon signed-rank test against Full for WAI; stratified permutation test for dependency recall).}
  \label{tab:ablation}
\end{table}

The components act on different properties of the dialogue rather than on its overall quality; basic scores barely change in any setting (4.92--4.94 vs.\ 4.95).
Removing the dependencies alone drops the dependency recall margin to chance (+0.3\,pp, n.s.), as low as removing the whole timeline (+0.1\,pp), so later sessions build on earlier events because the dependencies are given, not because events arrive in order.
Goal and task agreement, in contrast, fall less without dependencies ($-0.09$, $p<0.01$, and $-0.07$) but by 0.25 each once events are also shuffled ($p<0.01$): an agenda grounded in the student's week depends mainly on events arriving in a coherent order.
Removing the profile instead lowers the emotional bond ($-0.20$, $p<0.01$) and makes students more alike across trajectories (cross-student similarity 0.015 vs.\ 0.011), while dependency recall is preserved (+2.2\,pp).

\subsection{RQ4: Counseling Clients Whose Lives Change}
\label{sec:rq4}

\paragraph{Dynamic protocol.}
Existing multi-session evaluations hold the client's situation fixed between sessions. We instead let the client's life move on between sessions (Figure~\ref{fig:dynamic}). We generate 50 held-out students with the same pipeline and new random seeds, ten per trajectory archetype, and counsel each for ten weekly sessions. Each week, a simulated client (DeepSeek-V4-Flash with the student prompt of Appendix~\ref{app:prompt-student}) brings the week's event, while the counselor model sees only a de-identified intake record, the memory, the plan, and the dialogue. Memory and plans are updated by a shared module, so that models differ only in what the counselor says. Each model conducts 500 sessions.

\begin{figure}[t]
  \centering
  \resizebox{\columnwidth}{!}{% Figure: one week of the dynamic protocol (TikZ). \input from acl_latex.tex
\begin{tikzpicture}[
  font=\sffamily\scriptsize, >={Stealth[length=3.5pt]},
  st/.style={draw=black!45, fill=white, rounded corners=2pt, text width=1.66cm, minimum height=0.8cm, align=center, inner sep=1.5pt, font=\sffamily\tiny},
  lane/.style={font=\sffamily\tiny\bfseries, black!55, rotate=90, anchor=south},
  ar/.style={->, draw=black!60, line width=0.6pt}]
\def\ca{1.35}\def\cb{3.45}\def\cc{5.55}\def\cd{7.65}
\def\ya{3.05}\def\yb{1.85}\def\yf{0.25}
% lanes
\fill[black!3] (0.3,\ya-0.48) rectangle (8.6,\ya+0.48);
\node[lane] at (0.35,\ya) {World};
\node[lane] at (0.35,\yb) {Client};
\node[lane] at (0.35,\yf) {Counselor};
% steps
\node[st] (ev)  at (\ca,\ya) {\textbf{Event of week $t$}\\[-1pt]content, stress,\\[-1pt]dependencies};
\node[st] (cr)  at (\ca,\yb) {\textbf{Client reads}\\[-1pt]profile, event,\\[-1pt]antecedents, memory};
\node[st] (pre) at (\cb,\yb) {\textbf{PANAS}\\[-1pt]before session};
\node[st] (post) at (\cd,\yb) {\textbf{PANAS}\\[-1pt]after session};
\node[st] (kr)  at (\ca,\yf) {\textbf{Counselor reads}\\[-1pt]intake record,\\[-1pt]memory, plan};
\node[st] (up)  at (\cd,\yf) {\textbf{Shared module}\\[-1pt]updates memory,\\[-1pt]plans week $t{+}1$};
\node[draw=blue!55!black, fill=blue!7, rounded corners=2pt, line width=0.8pt, text width=1.66cm, minimum height=2.4cm, align=center, font=\sffamily\tiny] (se) at (\cc,{(\yb+\yf)/2}) {\textbf{Session}\\[-1pt]model under test\\[-1pt]${\le}\,24$ turns};
% flow
\draw[ar] (ev) -- (cr);
\draw[ar] (cr) -- (pre);
\draw[ar] (pre) -- (pre-|se.west);
\draw[ar] (post-|se.east) -- (post);
\draw[ar] (kr) -- (kr-|se.west);
\draw[ar] (up-|se.east) -- (up);
\draw[ar] (up.south) -- ++(0,-0.3) -| node[pos=0.25, below, font=\sffamily\tiny, black!65]{next week} (kr.south);
% barrier
\draw[black!55, dashed, line width=0.6pt] (0.3,1.15) -- (4.55,1.15);
\node[font=\sffamily\tiny\itshape, black!65, anchor=north, align=center, inner sep=1.5pt] at (2.55,1.15) {counselor never sees the event graph,\\[-1pt]core conflict, or stress level};
\end{tikzpicture}}
  \caption{One week of the dynamic protocol. The simulated client knows the week's event and its antecedents; the counselor model learns them only through the session. A session ends when the client says goodbye or after 24 turns.}
  \label{fig:dynamic}
\end{figure}

\paragraph{Metrics.}
Two pre-specified primary metrics capture whether the counselor follows a changing client: \emph{dependency recall} (D1), whether the counselor connects the week's event to the earlier events it depends on, and \emph{stress responsiveness} (D2), whether its pace follows the client's stress. Secondary metrics cover follow-up on the week's event (D3), fabricated references to earlier sessions (D4), the change in the client's affect within a session (D5), and session quality under the metrics of Section~\ref{sec:setup} (D6). Definitions are given in Appendix~\ref{app:dynamic}.

\begin{table*}[t]
  \centering
  \small
  \setlength{\tabcolsep}{4.5pt}
  \begin{tabular}{lccccccc}
    \toprule
    & \textbf{D1 dep.\ recall} & \textbf{D2 stress} & \textbf{D3 uptake} & \textbf{D3 proactive} & \textbf{D4 fabric.} & \textbf{D6 basic} & \textbf{D6 WAI} \\
    \textbf{Model} & (pp) $\uparrow$ & ($\rho$) $\uparrow$ & $\uparrow$ & $\uparrow$ & (\%) $\downarrow$ & $\uparrow$ & $\uparrow$ \\
    \midrule
    Base & +0.4 & 0.10 & 0.81 & 0.34 & 5.8 & 4.55 & 3.80 \\
    SMILE & +0.3 & 0.06 & 0.80 & 0.29 & 7.5 & 4.31 & 3.70 \\
    CPsyCoun & +0.3 & 0.12 & 0.83 & 0.31 & 6.8 & 4.46 & 3.77 \\
    MusPsy & +0.5 & 0.17 & 0.86 & 0.52 & 3.9 & 4.66 & 4.03 \\
    w/o timeline & +0.5 & 0.29 & 0.90 & 0.63 & 3.6 & 4.68 & 3.96 \\
    w/o dependencies & +0.6 & 0.30 & 0.90 & 0.62 & 3.4 & 4.69 & 4.00 \\
    w/o profile & +1.0 & 0.27 & 0.89 & 0.64 & 3.1 & 4.66 & 3.92 \\
    \textbf{CPCD-Chat} & \textbf{+1.1} & \textbf{0.31} & \textbf{0.91} & \textbf{0.66} & \textbf{2.9} & \textbf{4.70} & \textbf{4.05} \\
    \bottomrule
  \end{tabular}
  \caption{Dynamic protocol: counseling held-out students whose lives change between ten weekly sessions (500 sessions per model, shared-memory mode). All fine-tuned models share the same base model and differ only in their training data. CPCD-Chat, w/o dependencies, and w/o timeline average three training seeds (standard deviation at most 0.15\,pp for D1 and 0.02 for D6 basic).}
  \label{tab:dynamic}
\end{table*}

\begin{figure}[t]
  \centering
  \includegraphics[width=\columnwidth]{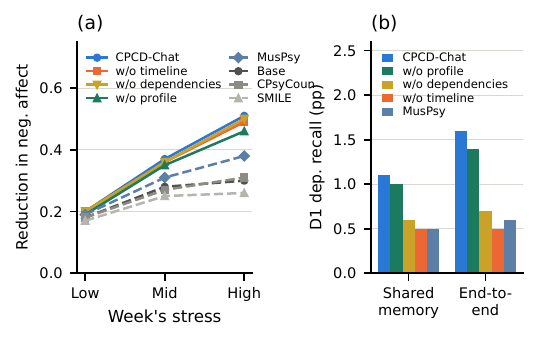}
  \caption{(a) Reduction in the simulated client's negative affect within a session (D5), by the stress level of the week. Solid lines: models trained on versions of CPCD; dashed lines: the other models. (b) Dependency recall (D1) when memory and plans come from a shared module and when each model maintains its own (end-to-end).}
  \label{fig:rq4}
\end{figure}

\paragraph{Results.}
Table~\ref{tab:dynamic} reports the results.
CPCD-Chat connects the week's event to its antecedents more than every model trained without dependencies or on other corpora (D1 +1.1\,pp, against +0.3--+0.6\,pp, and +0.4\,pp for Base); w/o profile is comparable (+1.0\,pp).
The dependencies in the training data account for this ability: removing them alone lowers D1 to +0.6\,pp, close to removing the whole timeline (+0.5\,pp), whereas the model trained without the profile keeps +1.0\,pp.
CPCD-Chat also adapts its pace to the client's stress (D2 $\rho=$ 0.31, against 0.10 for Base and at most 0.12 for the external baselines). Both timeline ablations retain this ability (0.30 and 0.29), as their training sessions still pair each event's stress level with a matching pace.
The MusPsy model, also trained on multi-session data with memory and plans, follows up on the week's event (0.86) and rarely fabricates (3.9\%), but its D1 (+0.5\,pp) and D2 (0.17) stay low: under our protocol, multi-session training without dependencies between events does not yield this ability.
Among the secondary metrics, CPCD-Chat asks about the client's week unprompted in 66\% of sessions, against 34\% for Base, and has the lowest fabrication rate among trained models (2.9\%).
Within sessions, the models trained on versions of CPCD reduce the client's negative affect more than the other models, and the gap grows with the week's stress level (Figure~\ref{fig:rq4}a): it is negligible in low-stress weeks and reaches 0.15--0.25 in high-stress weeks, with both timeline ablations close to CPCD-Chat.
CPCD-Chat's session quality also exceeds Base and the external baselines (D6 basic 4.70 vs.\ 4.31--4.55).
When each model instead maintains its own memory and plans its own sessions, the dependency-recall gap between CPCD-Chat and w/o dependencies widens from 0.5 to 0.9\,pp (Figure~\ref{fig:rq4}b; Appendix~\ref{app:dynamic}).

\subsection{RQ5: General Counseling Quality}
\label{sec:rq5}

\paragraph{Static protocol.}
To compare models under the setting of prior work, we counsel the same 50 held-out students for six sessions each with the profile only, so that nothing changes between sessions. The simulated client is the same as in the dynamic protocol and differs only in that no events advance. Sessions are rated with the quality metrics, and after each session the client completes the Positive and Negative Affect Schedule (PANAS)~\citep{watson1988panas}; we report its change from the first to the sixth session.

\begin{table}[t]
  \centering
  \small
  \setlength{\tabcolsep}{4pt}
  \begin{tabular}{lcccc}
    \toprule
    \textbf{Model} & \textbf{Basic} & \textbf{WAI} & \textbf{$\Delta$PA}$\uparrow$ & \textbf{$\Delta$NA}$\downarrow$ \\
    \midrule
    Base & 4.58 & 3.82 & +0.35 & $-$0.30 \\
    SMILE & 4.35 & 3.72 & +0.38 & $-$0.32 \\
    CPsyCoun & 4.48 & 3.80 & +0.42 & $-$0.36 \\
    MusPsy & 4.71 & \textbf{4.06} & +0.52 & $-$0.45 \\
    w/o timeline & 4.71 & 4.00 & +0.53 & $-$0.46 \\
    w/o dependencies & \textbf{4.72} & 4.01 & +0.54 & $-$0.47 \\
    w/o profile & 4.68 & 3.93 & +0.50 & $-$0.44 \\
    \textbf{CPCD-Chat} & \textbf{4.72} & 4.02 & \textbf{+0.55} & \textbf{$-$0.48} \\
    \bottomrule
  \end{tabular}
  \caption{Static protocol: six sessions per held-out student with no events between sessions. $\Delta$PA and $\Delta$NA are the changes in positive and negative affect from the first to the sixth session.}
  \label{tab:static}
\end{table}

\begin{table}[t]
  \centering
  \small
  \setlength{\tabcolsep}{3.5pt}
  \begin{tabular}{llcc}
    \toprule
    \textbf{Contrast} & \textbf{Metric} & \textbf{Static} & \textbf{Dynamic} \\
    \midrule
    \multirow{4}{*}{Dependencies} & D1 dep.\ recall & --- & +0.5$^{**}$ \\
     & D2 stress & --- & +0.01 \\
     & Basic & +0.00$^{\approx}$ & +0.01 \\
     & WAI & +0.01$^{\approx}$ & +0.05 \\

    \midrule
    \multirow{2}{*}{Profile} & D1 dep.\ recall & --- & +0.1 \\
     & WAI bond & +0.12$^{**}$ & +0.14$^{**}$ \\
    \midrule
    \multirow{4}{*}{MusPsy} & D1 dep.\ recall & --- & +0.6$^{**}$ \\
     & D2 stress & --- & +0.14$^{**}$ \\
     & Basic & +0.01$^{\approx}$ & +0.04 \\
     & WAI & $-$0.04$^{\approx}$ & +0.02 \\
    \bottomrule
  \end{tabular}
  \caption{Paired differences of CPCD-Chat against w/o dependencies (Dependencies), w/o profile (Profile), and the model trained on MusPsy (MusPsy) under the two protocols. $^{\approx}$: equivalent within $\pm0.15$ (TOST, $p<0.01$); $^{*}$: $p<0.05$; $^{**}$: $p<0.01$.}
  \label{tab:cross}
\end{table}

\paragraph{Results.}
Under the static protocol (Table~\ref{tab:static}), CPCD-Chat improves on Base and on both single-session baselines in session quality and in the client's affect over six sessions, and is on par with the model trained on MusPsy in basic quality (4.72 vs.\ 4.71). The MusPsy model reaches the highest working alliance (4.06), as its sessions are conditioned on counseling goals from case reports; CPCD-Chat trails it by 0.04, within the equivalence margin. Following a changing client thus costs no general counseling quality.

\paragraph{Protocol cross-check.}
Table~\ref{tab:cross} contrasts the components across protocols. The dependencies matter only when the client's life changes: CPCD-Chat and w/o dependencies are equivalent under the static protocol in both basic quality and working alliance, but differ under the dynamic protocol in dependency recall (+0.5\,pp, 95\% CI [+0.2, +0.8]). Shuffling the events as well adds little to this gap (+0.6\,pp) but lowers the working alliance (+0.09). The profile's effect on the emotional bond, in contrast, appears under both protocols, and the MusPsy model shows a similar pattern but also trails in stress responsiveness (+0.14).

\paragraph{Robustness and human evaluation.}
With a second simulated client built on a different LLM, the dependency-recall gap between CPCD-Chat and w/o dependencies shrinks but persists (+0.4\,pp, 95\% CI [+0.1, +0.8]). The same experts compared 50 blind pairs of CPCD-Chat and w/o dependencies sessions for weeks whose event has an antecedent. CPCD-Chat connected the event correctly to an earlier one in 58\% of sessions against 34\%, raising it before the client in 34\% against 12\%, and was preferred in 31 of 44 non-tied pairs (sign test $p=$ 0.01). Correctly connected sessions have higher D1 ($r=$ 0.41; Appendix~\ref{app:human}).

% ==================================================================
\section{Conclusion}
% ==================================================================

We presented Psy-Chronicle, which grounds multi-session counseling synthesis in a student's semester timeline, and used it to build CPCD, an open Chinese dataset of semester-long campus counseling.
The generated dialogues follow their timelines: dependencies between events carry across sessions, and references to earlier sessions rarely introduce content absent from the trajectory.
Under a published evaluation protocol, CPCD matches the multi-session corpus MusPsy on basic counseling quality and exceeds all single-session corpora on the working alliance, and the event graph and the profile shape different properties of the dialogue.
A model fine-tuned on CPCD connects each week's events to earlier ones more often than a model trained on the same data without dependencies between events, and the two are equivalent when the client's situation stays fixed.

\section*{Limitations}
\paragraph{Realism of synthetic data.}
CPCD is generated by LLMs and cannot fully reproduce real counseling, such as silence, resistance, or nonlinear change. Real campus counseling records are subject to strict ethical, privacy, and access constraints that limit their collection and sharing at scale, so we regard synthesis as a necessary compromise. Our consistency checks show that the dialogues follow the profiles and timelines they were conditioned on; they do not show that these profiles and timelines are clinically representative of real students.

\paragraph{LLM-based evaluation.}
Our evaluation relies on an LLM judge following a published protocol, which carries known biases. The WAI-SR is completed by the judge on behalf of a simulated client rather than by a real one, and several of its items presuppose a relationship built over earlier sessions. Likewise, the PANAS is completed by a simulated client and the CTRS by an LLM rater. We therefore treat these scores as proxy ratings in a simulated setting rather than as measurements that inherit the validity of the original instruments. Expert ratings preserve the judge's ranking of corpora but agree with it only moderately on individual sessions (Appendix~\ref{app:human}).

\paragraph{Simulated evaluation.}
Models are evaluated with simulated clients whose profiles and event graphs come from the same pipeline as the training data, so the dynamic protocol favors models trained on CPCD over the external baselines; we therefore base our claims about the timeline on comparisons among models trained on versions of CPCD. The human evaluation involves two raters, 200 dataset sessions, and 50 session pairs from the dynamic protocol.

\paragraph{Simulation assumptions.}
Several rules of the simulator are design choices rather than empirical claims: personality types serve only to diversify the profiles, not as a validated psychometric construct; the five trajectory archetypes are chosen to cover distinct dynamics rather than estimated from student data; every background includes a risk factor from which the core conflict is derived; each event yields exactly one session; and the CBT stage advances by a stress-based heuristic and always ends in consolidation, regardless of whether goals are met. These rules shape what models trained on CPCD learn, and real students may present protective factors, situational difficulties without a single underlying conflict, or a different pace of counseling.

\paragraph{Fixed trajectories.}
Each event graph is fixed before counseling begins, including events such as seeking help or confiding in a roommate that real counseling might bring about. Counseling therefore cannot change later events. The dynamic protocol tests whether a model follows a client whose life changes, not whether it changes the course of that life, and the PANAS results describe the simulated client's reaction within a session (D5) or over six simulated sessions (static protocol) rather than longer-term outcomes.

\paragraph{Scope.}
This work studies how grounding synthesis in a client's background and evolving life shapes counseling data and the models trained on it; it does not address clinical deployment. Crisis intervention, including risk assessment and referral to support resources, is outside this scope: CPCD does not model these steps, and we do not evaluate them. Some health events nonetheless approach this boundary, for example episodes of disordered eating (Appendix~\ref{app:conformance}). All sessions follow a CBT structure, which may not suit every student or presenting problem. Like other synthetic counseling corpora, CPCD is a research resource for developing and analyzing counseling models rather than a clinical instrument. Our expert ratings assess session quality and cross-session integration, not clinical validity; the raters have professional experience in psychology but are not licensed clinicians, and review by licensed clinicians would be required before any clinical use.

\paragraph{Single generator.}
All dialogues are produced by one LLM in a single generation run. The same model also plays the student and the counselor, which may make the two sides of the dialogue more alike than in real counseling. The counselor shows recurring formulations, the most frequent of which appears in 8.5\% of counselor utterances (Appendix~\ref{app:conformance}), and we report no variance across repeated generation runs.

\section*{Ethical Considerations}
CPCD contains no real counseling records; all students are synthetic. The two expert raters gave informed consent and were paid for their time; they rated synthetic dialogues only. The dataset and CPCD-Chat are released for research only and are not intended for deployment as a counseling service or as a substitute for professional care. Models trained on CPCD have not been evaluated in crisis situations and should not be used with users who may be at risk. Sessions that contain suicide- or self-harm-related expressions are marked in the release (Appendix~\ref{app:conformance}).

\bibliography{custom}

@inproceedings{wang-etal-2026-psychological,
  title     = {Psychological Counseling Cannot Be Achieved Overnight: Automated Psychological Counseling Through Multi-Session Conversations},
  author    = {Wang, Bichen and Wang, Junzhe and Sun, Yixin and Fu, Xing and Zhao, Yanyan and Qin, Bing},
  booktitle = {Findings of the Association for Computational Linguistics: ACL 2026},
  pages     = {16593--16609},
  year      = {2026}
}

@inproceedings{qiu2025psydial,
  title     = {{PsyDial}: A Large-scale Long-term Conversational Dataset for Mental Health Support},
  author    = {Qiu, Huachuan and Lan, Zhenzhong},
  booktitle = {Proceedings of the 63rd Annual Meeting of the Association for Computational Linguistics (Volume 1: Long Papers)},
  pages     = {21624--21655},
  year      = {2025},
  doi       = {10.18653/v1/2025.acl-long.1049}
}

@article{zhou2025diacbt,
  title   = {{DiaCBT}: A Long-Periodic Dialogue Corpus Guided by Cognitive Conceptualization Diagram for {CBT}-based Psychological Counseling},
  author  = {Zhou, Yougen and Zhou, Ningning and Chen, Qin and Zhou, Jie and Zhou, Aimin and He, Liang},
  journal = {arXiv preprint arXiv:2509.02999},
  year    = {2025}
}

@article{liu2026dmtcbt,
  title   = {{DMT-CBT}: Longitudinal Therapeutic State Modeling for {CBT} Counseling},
  author  = {Liu, Chang and Zhang, Shuyi and Ma, Changsheng and Tao, Yongfeng and Yang, Minqiang and Hu, Bin},
  journal = {arXiv preprint arXiv:2606.03132},
  year    = {2026}
}

@article{li2026longcounsel,
  title   = {{LongCounsel-8}: A Benchmark Suite for Longitudinal Depression Tracking from Multi-Session Counseling Dialogues},
  author  = {Li, Jiayi and Wu, Zhaomin and He, Bingsheng},
  journal = {arXiv preprint arXiv:2609.03507},
  year    = {2026}
}

@inproceedings{zhao2026tudial,
  title     = {{TuDialCorpus}: A School Counseling Conversation Dataset with Hierarchical Stage Transition and Supportive Strategy},
  author    = {Zhao, Yinan and Guo, Ruiqiang and Jia, Xu},
  booktitle = {Proceedings of the International Conference on Intelligent Computing (ICIC), Lecture Notes in Computer Science, vol. 16665},
  pages     = {451--463},
  year      = {2026}
}

@article{auerbach2018who,
  title   = {{WHO} World Mental Health Surveys International College Student Project: Prevalence and Distribution of Mental Disorders},
  author  = {Auerbach, Randy P. and Mortier, Philippe and Bruffaerts, Ronny and Alonso, Jordi and Benjet, Corina and Cuijpers, Pim and Demyttenaere, Koen and Ebert, David D. and Green, Jennifer Greif and Hasking, Penelope and Murray, Elaine and Nock, Matthew K. and Pinder-Amaker, Stephanie and Sampson, Nancy A. and Stein, Dan J. and Vilagut, Gemma and Zaslavsky, Alan M. and Kessler, Ronald C.},
  journal = {Journal of Abnormal Psychology},
  volume  = {127},
  number  = {7},
  pages   = {623--638},
  year    = {2018},
  doi     = {10.1037/abn0000362}
}

@article{gao2020prevalence,
  title   = {Prevalence of Depression among {Chinese} University Students: A Systematic Review and Meta-Analysis},
  author  = {Gao, Lu and Xie, Yuanchen and Jia, Chunhua and Wang, Wei},
  journal = {Scientific Reports},
  volume  = {10},
  year    = {2020},
  doi     = {10.1038/s41598-020-72998-1}
}

@inproceedings{wang2014studentlife,
  title     = {{StudentLife}: Assessing Mental Health, Academic Performance and Behavioral Trends of College Students Using Smartphones},
  author    = {Wang, Rui and Chen, Fanglin and Chen, Zhenyu and Li, Tianxing and Harari, Gabriella and Tignor, Stefanie and Zhou, Xia and Ben-Zeev, Dror and Campbell, Andrew T.},
  booktitle = {Proceedings of the 2014 ACM International Joint Conference on Pervasive and Ubiquitous Computing (UbiComp)},
  pages     = {3--14},
  year      = {2014},
  doi       = {10.1145/2632048.2632054}
}

@misc{moe2023action,
  title        = {Special Action Plan for Comprehensively Strengthening and Improving Student Mental Health Work in the New Era (2023--2025)},
  note         = {In Chinese},
  author       = {{Ministry of Education of China}},
  year         = {2023},
  howpublished = {Policy document}
}

@book{beck2011cbt,
  title     = {Cognitive Behavior Therapy: Basics and Beyond},
  author    = {Beck, Judith S.},
  edition   = {2nd},
  publisher = {Guilford Press},
  year      = {2011}
}

@inproceedings{sun2021psyqa,
  title     = {{PsyQA}: A {Chinese} Dataset for Generating Long Counseling Text for Mental Health Support},
  author    = {Sun, Hao and Lin, Zhenru and Zheng, Chujie and Liu, Siyang and Huang, Minlie},
  booktitle = {Findings of the Association for Computational Linguistics: ACL-IJCNLP 2021},
  pages     = {1489--1503},
  year      = {2021}
}

@inproceedings{liu2021esconv,
  title     = {Towards Emotional Support Dialog Systems},
  author    = {Liu, Siyang and Zheng, Chujie and Demasi, Orianna and Sabour, Sahand and Li, Yu and Yu, Zhou and Jiang, Yong and Huang, Minlie},
  booktitle = {Proceedings of the 59th Annual Meeting of the Association for Computational Linguistics and the 11th International Joint Conference on Natural Language Processing (Volume 1: Long Papers)},
  pages     = {3469--3483},
  year      = {2021}
}

@inproceedings{qiu2024smile,
  title     = {{SMILE}: Single-turn to Multi-turn Inclusive Language Expansion via {ChatGPT} for Mental Health Support},
  author    = {Qiu, Huachuan and He, Hongliang and Zhang, Shuai and Li, Anqi and Lan, Zhenzhong},
  booktitle = {Findings of the Association for Computational Linguistics: EMNLP 2024},
  pages     = {615--636},
  year      = {2024}
}

@inproceedings{zhang2024cpsycoun,
  title     = {{CPsyCoun}: A Report-based Multi-turn Dialogue Reconstruction and Evaluation Framework for {Chinese} Psychological Counseling},
  author    = {Zhang, Chenhao and Li, Renhao and Tan, Minghuan and Yang, Min and Zhu, Jingwei and Yang, Di and Zhao, Jiahao and Ye, Guancheng and Li, Chengming and Hu, Xiping},
  booktitle = {Findings of the Association for Computational Linguistics: ACL 2024},
  pages     = {13947--13966},
  year      = {2024}
}

@inproceedings{lee2024cactus,
  title     = {{CACTUS}: Towards Psychological Counseling Conversations using Cognitive Behavioral Theory},
  author    = {Lee, Suyeon and Kim, Sunghwan and Kim, Minju and Kang, Dongjin and Yang, Dongil and Kim, Harim and Kang, Minseok and Jung, Dayi and Kim, Min Hee and Lee, Seungbeen and Chung, Kyong-Mee and Yu, Youngjae and Lee, Dongha and Yeo, Jinyoung},
  booktitle = {Findings of the Association for Computational Linguistics: EMNLP 2024},
  pages     = {14245--14274},
  year      = {2024}
}

@misc{young1980ctrs,
  title        = {Cognitive Therapy Scale: Rating Manual},
  author       = {Young, Jeffrey and Beck, Aaron T.},
  year         = {1980},
  howpublished = {Unpublished manuscript, Center for Cognitive Therapy, University of Pennsylvania}
}

@inproceedings{maharana2024locomo,
  title     = {Evaluating Very Long-Term Conversational Memory of {LLM} Agents},
  author    = {Maharana, Adyasha and Lee, Dong-Ho and Tulyakov, Sergey and Bansal, Mohit and Barbieri, Francesco and Fang, Yuwei},
  booktitle = {Proceedings of the 62nd Annual Meeting of the Association for Computational Linguistics (Volume 1: Long Papers)},
  pages     = {13851--13870},
  year      = {2024},
  doi       = {10.18653/v1/2024.acl-long.747}
}

@inproceedings{xu2022msc,
  title     = {Beyond Goldfish Memory: Long-Term Open-Domain Conversation},
  author    = {Xu, Jing and Szlam, Arthur and Weston, Jason},
  booktitle = {Proceedings of the 60th Annual Meeting of the Association for Computational Linguistics (Volume 1: Long Papers)},
  pages     = {5180--5197},
  year      = {2022}
}

@inproceedings{jang2023chronicles,
  title     = {Conversation Chronicles: Towards Diverse Temporal and Relational Dynamics in Multi-Session Conversations},
  author    = {Jang, Jihyoung and Boo, Minseong and Kim, Hyounghun},
  booktitle = {Proceedings of the 2023 Conference on Empirical Methods in Natural Language Processing},
  pages     = {13584--13606},
  year      = {2023}
}

@article{mandal2026graph2counsel,
  title   = {{Graph2Counsel}: Clinically Grounded Synthetic Counseling Dialogue Generation from Client Psychological Graphs},
  author  = {Mandal, Aishik and Arnaout, Hiba and Ong, Clarissa W. and Bockhorst, Juliet and Sheehan, Kate and Moldow, Rachael and Chakraborty, Tanmoy and Gurevych, Iryna},
  journal = {arXiv preprint arXiv:2604.20382},
  year    = {2026}
}

@article{shi2025cpsdd,
  title   = {Toward Real-World {Chinese} Psychological Support Dialogues: {CPsDD} Dataset and a Co-Evolving Multi-Agent System},
  author  = {Shi, Yuanchen and Zhang, Longyin and Kong, Fang},
  journal = {arXiv preprint arXiv:2507.07509},
  year    = {2025}
}

@article{liu2023chatcounselor,
  title   = {{ChatCounselor}: A Large Language Models for Mental Health Support},
  author  = {Liu, June M. and Li, Donghao and Cao, He and Ren, Tianhe and Liao, Zeyi and Wu, Jiamin},
  journal = {arXiv preprint arXiv:2309.15461},
  year    = {2023}
}

@article{hu2025psyllm,
  title   = {Beyond Empathy: Integrating Diagnostic and Therapeutic Reasoning with Large Language Models for Mental Health Counseling},
  author  = {Hu, He and Zhou, Yucheng and Si, Juzheng and Wang, Qianning and Zhang, Hengheng and Ren, Fuji and Ma, Fei and Cui, Laizhong and Tian, Qi},
  journal = {arXiv preprint arXiv:2505.15715},
  year    = {2025}
}

@article{hatcher2006waisr,
  title   = {Development and Validation of a Revised Short Version of the {Working Alliance Inventory}},
  author  = {Hatcher, Robert L. and Gillaspy, J. Arthur},
  journal = {Psychotherapy Research},
  volume  = {16},
  number  = {1},
  pages   = {12--25},
  year    = {2006},
  doi     = {10.1080/10503300500352500}
}

@article{horvath1989wai,
  title   = {Development and Validation of the {Working Alliance Inventory}},
  author  = {Horvath, Adam O. and Greenberg, Leslie S.},
  journal = {Journal of Counseling Psychology},
  volume  = {36},
  number  = {2},
  pages   = {223--233},
  year    = {1989},
  doi     = {10.1037/0022-0167.36.2.223}
}

@article{munder2010waisr,
  title   = {Working Alliance Inventory-Short Revised ({WAI-SR}): Psychometric Properties in Outpatients and Inpatients},
  author  = {Munder, Thomas and Wilmers, Fabian and Leonhart, Rainer and Linster, Hans Wolfgang and Barth, J{\"u}rgen},
  journal = {Clinical Psychology \& Psychotherapy},
  volume  = {17},
  number  = {3},
  pages   = {231--239},
  year    = {2010},
  doi     = {10.1002/cpp.658}
}

@article{horvath1991alliance,
  title   = {Relation between Working Alliance and Outcome in Psychotherapy: A Meta-Analysis},
  author  = {Horvath, Adam O. and Symonds, B. Dianne},
  journal = {Journal of Counseling Psychology},
  volume  = {38},
  number  = {2},
  pages   = {139--149},
  year    = {1991},
  doi     = {10.1037/0022-0167.38.2.139}
}

@article{horvath2011alliance,
  title   = {Alliance in Individual Psychotherapy},
  author  = {Horvath, Adam O. and Del Re, A. C. and Fl{\"u}ckiger, Christoph and Symonds, Dianne},
  journal = {Psychotherapy},
  volume  = {48},
  number  = {1},
  pages   = {9--16},
  year    = {2011},
  doi     = {10.1037/a0022186}
}

@article{qwen2025qwen25,
  title   = {{Qwen2.5} Technical Report},
  author  = {{Qwen Team}},
  journal = {arXiv preprint arXiv:2412.15115},
  year    = {2024}
}

@inproceedings{hu2022lora,
  title     = {{LoRA}: Low-Rank Adaptation of Large Language Models},
  author    = {Hu, Edward J. and Shen, Yelong and Wallis, Phillip and Allen-Zhu, Zeyuan and Li, Yuanzhi and Wang, Shean and Wang, Lu and Chen, Weizhu},
  booktitle = {International Conference on Learning Representations},
  year      = {2022}
}

@inproceedings{zheng2024llamafactory,
  title     = {{LlamaFactory}: Unified Efficient Fine-Tuning of 100+ Language Models},
  author    = {Zheng, Yaowei and Zhang, Richong and Zhang, Junhao and Ye, Yanhan and Luo, Zheyan},
  booktitle = {Proceedings of the 62nd Annual Meeting of the Association for Computational Linguistics (Volume 3: System Demonstrations)},
  pages     = {400--410},
  year      = {2024}
}

@article{schuirmann1987tost,
  title   = {A Comparison of the Two One-Sided Tests Procedure and the Power Approach for Assessing the Equivalence of Average Bioavailability},
  author  = {Schuirmann, Donald J.},
  journal = {Journal of Pharmacokinetics and Biopharmaceutics},
  volume  = {15},
  number  = {6},
  pages   = {657--680},
  year    = {1987},
  doi     = {10.1007/BF01068419}
}

@article{watson1988panas,
  title   = {Development and Validation of Brief Measures of Positive and Negative Affect: The {PANAS} Scales},
  author  = {Watson, David and Clark, Lee Anna and Tellegen, Auke},
  journal = {Journal of Personality and Social Psychology},
  volume  = {54},
  number  = {6},
  pages   = {1063--1070},
  year    = {1988}
}

\appendix

% ==================================================================
\section{Student Profiles}
\label{app:profile}
% ==================================================================

\begin{figure*}[t]
  \centering
  \includegraphics[width=\textwidth]{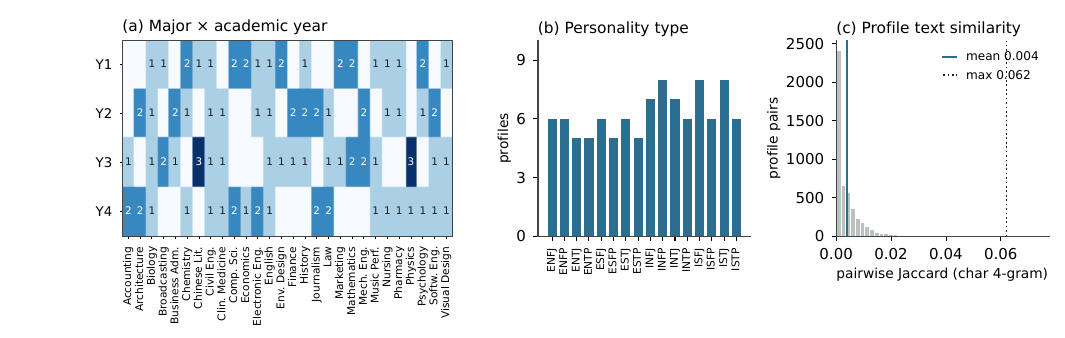}
  \caption{Attribute coverage and textual diversity of the 100 student profiles. (a) Joint distribution of major and academic year; cells give profile counts. (b) Distribution over the 16 personality types. (c) Pairwise Jaccard similarity between profiles over character 4-grams of the concatenated \code{background} and \code{core\_conflict} fields (4{,}950 pairs).}
  \label{fig:profile}
\end{figure*}

Each profile records demographics (gender, age, academic year, major), personality (type, traits, coping style), family and social background, a core psychological conflict, and a primary stress domain.

\paragraph{Factorial attribute design.}
We enumerate the $2\times4\times5=40$ cells of gender, academic year, and stress domain, and allocate the 100 profiles as evenly as divisibility permits, distributing the remainder over randomly chosen cells. Majors are drawn from an independently permuted pool in which each of the 28 majors appears three or four times, so major is uncorrelated with the factorial block by construction (Figure~\ref{fig:profile}a).

\paragraph{Controlled identity attributes.}
Personality type and name are fixed programmatically, since the generator otherwise concentrates both on a few frequent values. Personality types are sampled over all 16 MBTI types, reserving one slot per type before distributing the remainder with a mild weighting toward introverted types (Figure~\ref{fig:profile}b). Names are drawn without replacement from a gender-conditioned pool. All controlled attributes are stated in the prompt as hard constraints and overwritten in the returned JSON.

\paragraph{Redundancy control.}
For each new profile we compute the character 4-gram Jaccard similarity of its \code{background} and \code{core\_conflict} fields against every accepted profile. A profile above $0.30$ is regenerated with an instruction to avoid common configurations, for at most two rounds. No released pair exceeds $0.062$ (Figure~\ref{fig:profile}c).

\paragraph{Resulting profiles.}
Table~\ref{tab:profile} summarizes the released profiles. No major is confined to a single year. Five majors appear under a single gender, which matches the expectation under independence (5.0); we do not force balance, as doing so would reintroduce a dependence between major and gender.

\begin{table}[t]
  \centering\footnotesize
  \begin{tabularx}{\columnwidth}{@{}Xr@{}}
    \toprule
    Gender (male\,:\,female)                & 49\,:\,51 \\
    Academic year (Y1/Y2/Y3/Y4)             & 25/24/26/25 \\
    Profiles per stress domain              & 19--22 \\
    Majors covered / profiles per major     & 28 of 28 / 3--4 \\
    Personality types / profiles per type   & 16 of 16 / 5--8 \\
    Distinct names                          & 100 \\
    \midrule
    Majors realized under one year only     & 0 / 28 \\
    Majors realized under one gender only   & 5 / 28 \\
    Distinct (gender, year, major, domain)  & 97 / 100 \\
    \midrule
    Pairwise similarity, mean / max         & 0.004 / 0.062 \\
    \code{background} length, mean (chars)  & 331 \\
    Distinct coping-style descriptors       & 37 \\
    \bottomrule
  \end{tabularx}
  \caption{Composition of the 100 student profiles.}
  \label{tab:profile}
\end{table}

% ==================================================================
\section{Temporal Stress Event Graphs}
\label{app:event}
% ==================================================================

\begin{figure*}[t]
  \centering
  \includegraphics[width=\textwidth]{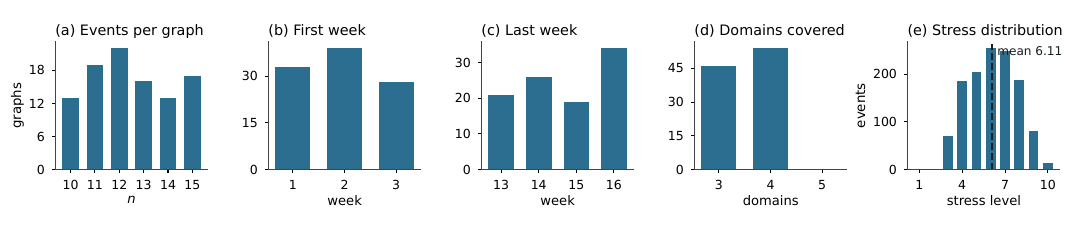}
  \caption{Structure of the 100 event graphs. (a)--(d) Distributions of the sampled specification: event count, week of the first and last event, and number of stress domains covered. (e) Stress levels over all 1{,}248 events.}
  \label{fig:structure}
\end{figure*}

Each node of an event graph records the event content, week, stress domain, stress level (1--10), psychological impact, and a list \code{depends\_on} of earlier events.

\paragraph{Specification sampling.}
Structural properties are sampled before generation and stated as exact requirements, because a generator given a range tends to realize its endpoint. For each student we draw the event count $n\in\{10,\dots,15\}$, the first week in $\{1,2,3\}$, the last week in $\{13,\dots,16\}$, and a trajectory archetype uniformly from the five in Table~\ref{tab:shapes}, each passed to the generator with a one-sentence description. The profile's primary domain receives $\max(4,\lceil 0.45n\rceil)$ events and two or three secondary domains share the rest, so each trajectory has its own focus while the corpus stays balanced (241--260 events per domain).

\paragraph{Archetype-conditioned stress constraints.}
Each archetype carries its own constraint on low-stress events (Table~\ref{tab:shapes}), with one universal floor: no trajectory consists entirely of events above level~5. Across the corpus the mean stress level is 6.11, with 20.5\% of events at or below level~4 and 22.8\% at or above level~8 (Figure~\ref{fig:structure}e), so trajectories contain periods of relief as well as peaks.

\begin{table}[t]
  \centering\footnotesize
  \begin{tabularx}{\columnwidth}{@{}lX@{}}
    \toprule
    \textbf{Archetype} & \textbf{Constraint} \\
    \midrule
    Escalating    & $n_{\le 5}\ge 2$; first event at level $\le 6$ \\
    Spike--relief & $n_{\le 4}\ge 2$ and $n_{\le 5}\ge 3$ \\
    Oscillating   & $n_{\le 4}\ge 2$; reversals $\ge 3$ \\
    Late onset    & mean level of the first $\lceil n/3\rceil$ events $\le 5.8$; $n_{\le 5}\ge 3$ \\
    Early crisis  & $n_{\le 5}\ge 3$, of which $n_{\le 4}\ge 1$ \\
    \midrule
    \emph{all}    & $n_{\le 5}\ge 1$ \\
    \bottomrule
  \end{tabularx}
  \caption{Archetype-conditioned stress constraints. $n_{\le k}$ is the number of events at stress level at most $k$; reversals count sign changes in the stress sequence.}
  \label{tab:shapes}
\end{table}

\paragraph{Validation and revision.}
Every graph is checked against twelve predicates in four groups. \emph{Schema}: all fields present, stress level an integer in $[1,10]$, domain among the five permitted values. \emph{Structure}: node count equal to $n$; identifiers $E_1,\dots,E_n$ unique; weeks integral, non-decreasing, and spanning the sampled first and last week; per-domain counts within one of quota. \emph{Dependency}: every \code{depends\_on} reference resolves to an existing node in a strictly earlier week, the graph is acyclic, and at least half of the non-initial nodes have a predecessor. \emph{Trajectory}: the constraints of Table~\ref{tab:shapes}. Violations are rendered in natural language and returned to the generator for up to three rounds. A graph still failing is passed through a mechanical repair that sorts nodes by week, renumbers identifiers, and removes invalid references, without editing content. All 100 released graphs satisfy all twelve predicates.

\paragraph{Resulting graphs.}
Table~\ref{tab:event} summarizes the released graphs. Most events build on earlier ones while roots remain, giving traceable chains of mean length 8.6 rather than a saturated structure in which nearly every event depends on all earlier ones.

\begin{table}[t]
  \centering\footnotesize
  \begin{tabularx}{\columnwidth}{@{}Xr@{}}
    \toprule
    Graphs / events                       & 100 / 1{,}248 \\
    Events per graph, mean (range)        & 12.5 (10--15) \\
    First / last week                     & 1--3 / 13--16 \\
    Stress domains per graph              & 3--4 \\
    \midrule
    Stress level, mean (median)           & 6.11 (6) \\
    \quad events at level $\le 4$ / $\ge 8$ & 20.5\% / 22.8\% \\
    \midrule
    Dependency edges                      & 1{,}735 \\
    \quad events with $\ge 1$ / $\ge 2$ predecessors & 90.1\% / 44.3\% \\
    Longest dependency chain, mean        & 8.6 \\
    \midrule
    Event-content similarity, mean / max  & 0.005 / 0.024 \\
    \bottomrule
  \end{tabularx}
  \caption{Statistics of the 100 temporal stress event graphs. Similarity is the pairwise Jaccard over character 4-grams of concatenated event descriptions across students.}
  \label{tab:event}
\end{table}

\paragraph{Alignment with the academic calendar.}
The generation prompt specifies the first and last week and the trajectory archetype, but no exam schedule. Of the 111 events that mention midterms, 58.6\% fall in weeks 7--9, peaking in week 8; of the 129 that mention final exams, 76.0\% fall in weeks 13--16.

% ==================================================================
\section{Case Study}
\label{app:case}
% ==================================================================

We trace one student through the semester to show how the profile, the event graph, and the memory shape the dialogue. The student is the one whose correlations of stress with session length and with student affect ($r=0.60$ and $0.19$) lie closest to the corpus medians ($0.68$ and $0.16$) among students whose sessions are all complete and well formed. Excerpts are translated from Chinese.

The student is a male first-year computer-science student whose primary stress domain is interpersonal relationships. His core conflict, visible only to the student agent, is a longing for genuine connection undermined by a habit of suppressing his own needs to placate others, formed between a stern father and a silent mother. His event graph follows the spike--relief archetype over 12 sessions (Table~\ref{tab:case}).

\begin{table*}[t]
  \centering\footnotesize
  \setlength{\tabcolsep}{4pt}
  \begin{tabularx}{\textwidth}{@{}rrXlrrll@{}}
    \toprule
    \textbf{S} & \textbf{Week} & \textbf{Event (summary)} & \textbf{Domain} & \textbf{Stress} & \textbf{Turns} & \textbf{Stage} & \textbf{Depends on} \\
    \midrule
    1  & 2  & Silent at the first dorm night talk; over-apologizes when teased & Interp. & 6 & 19 & 1 & --- \\
    2  & 3  & TA in the lab: ``A CS student can't even handle this?'' & Academic & 5 & 22 & 1 & S1 \\
    3  & 4  & Says ``it's fine'' when a roommate breaks his charger & Interp. & 6 & 20 & 2 & S1 \\
    4  & 5  & Week of insomnia; goes blank when called on in calculus & Health & 5 & 21 & 2 & S2, S3 \\
    \textbf{5} & \textbf{6} & \textbf{Left out at the class barbecue; rides back alone} & \textbf{Interp.} & \textbf{8} & \textbf{23} & \textbf{2 (hold)} & \textbf{S1, S3} \\
    6  & 7  & Leaves the last midterm question blank; bottom 20\% & Academic & 7 & 23 & 3 & S2, S5 \\
    \textbf{7} & \textbf{8} & \textbf{Acute anxiety attack in the canteen} & \textbf{Health} & \textbf{9} & \textbf{24} & \textbf{3 (hold)} & \textbf{S4--S6} \\
    8  & 9  & Cries and first discloses childhood name-calling & Interp. & 4 & 22 & 5 (acc.) & S7 \\
    9  & 10 & Asks the TA for help; the TA once failed a course too & Academic & 4 & 22 & 6 (acc.) & S6, S8 \\
    10 & 11 & Tells roommates about the name-calling; nobody laughs & Interp. & 3 & 21 & 6 (acc.) & S8, S9 \\
    11 & 12 & Runs with a roommate: ``My dad is like that too'' & Health & 3 & 18 & 6 (acc.) & S8, S10 \\
    12 & 14 & Proposes his own plan in a group meeting; texts his father & Interp. & 5 & 21 & 6 & S9--S11 \\
    \bottomrule
  \end{tabularx}
  \caption{Semester trajectory of the case student. Turns counts student turns. Stage is the CBT stage of Table~\ref{tab:stages}; \emph{hold} keeps the stage at a stress peak and \emph{acc.}\ advances faster during relief. Depends on lists the dependency edges of the event graph. Bold rows mark the stress peaks.}
  \label{tab:case}
\end{table*}

\paragraph{Event-driven pacing.}
Stress peaks at S5 and S7 (levels 8 and 9) and falls to 3--4 in S8--S11. Session length follows: 24 student turns at the level-9 peak against 18--21 at level 3. The stage is held at both peaks and advances quickly once stress falls, from stage 3 to stages 5 and 6 within two sessions.

\paragraph{Dependency recall.}
The graph links the TA's remark (S2) to the blank-out in calculus two weeks later (S4). In S4 the link surfaces in the dialogue, and the student reproduces the remark verbatim.

\begin{casebox}{Excerpt 1 (S4, week 5)}
\textbf{Counselor:} \dots In those seconds you were standing, was there a moment you thought of what the TA said in the lab?\par
\textbf{Student:} Kind of\dots\ The moment the teacher called my name my mind went blank, and then the TA's line popped up: ``A CS student can't even handle this.'' Calculus has nothing to do with it, but I couldn't stop it.\par
\textbf{Counselor:} So that line doesn't stay in the lab room.
\end{casebox}

\paragraph{Gradual disclosure of the core conflict.}
The counselor never sees the core conflict, and no family member is mentioned in S1--S3. It first emerges in S4, when the student answers an open question in his own words.

\begin{casebox}{Excerpt 2 (S4, week 5)}
\textbf{Counselor:} From the dorm talk to the TA to the broken charger\dots\ you seem to measure yourself with the same ruler. Who gave you this ruler?\par
\textbf{Student:} I never really thought about that\dots\ It's been like this since I was little. My dad\dots\ he doesn't mean harm, he just thinks boys shouldn't be soft: you fall, you get up on your own. My mom doesn't really get involved.
\end{casebox}

\noindent The theme deepens in S6 (``If I did badly, my dad would just say, `What's there to cry about?'\,''), and in S12 the student takes a first step toward it by texting his father.

\paragraph{Memory across sessions.}
In the final session the counselor recalls a sentence from S1, and the student reconstructs his progress from three earlier sessions (S9--S11).

\begin{casebox}{Excerpt 3 (S12, week 14)}
\textbf{Counselor:} Early this term you told me, ``Apologizing beats making people unhappy.'' This week your voice shook, but you finished presenting your plan. Looking back, what brought you here?\par
\textbf{Student:} That time you had me ask the TA, I was ready to say ``Sorry, I'm too dumb,'' but he said he had failed a course in his first year too. Then, while running, Zhang Lei said ``My dad is like that too.'' And last time at the night talk, I told them about the nickname and nobody laughed. It added up bit by bit.
\end{casebox}

\noindent In this case, the event graph sets what each session is about and how intense it is, the profile surfaces only through the student, and the memory lets both agents build on earlier sessions.

% ==================================================================
\section{Dialogue Conformance}
\label{app:conformance}
% ==================================================================

The dialogues are generated from profiles and event graphs that are fixed in advance. Each check below compares a property of these inputs with a property measured on the generated text, so the expected direction is set before the text exists. Where a check could be satisfied by chance, we report a permutation null with 1{,}000 resamples. C1 and C2 shuffle stress levels across all 1{,}248 sessions. C4 shuffles the dependency label among all session pairs with the same gap, pooling students, and uses only earlier events with at least eight anchor terms and gaps with at least 15 linked and 15 unlinked pairs. These nulls do not model correlation among sessions of the same student; the cross-student pairing in C4 serves as a complementary baseline. C1 and C3 verify mechanisms that the pipeline sets directly and C6 verifies an input restriction, so they are sanity checks of the pipeline rather than evidence about the dialogues. These checks establish that the generated dialogues follow their specification; they do not establish clinical validity.

\begin{table}[t]
  \centering\footnotesize
  \begin{tabular}{@{}lr@{}}
    \toprule
    Students / sessions            & 100 / 1{,}248 \\
    Utterances                     & 54{,}176 \\
    Characters                     & 4{,}252{,}810 \\
    Sessions per student           & 12.48 (10--15) \\
    Turns per session              & 21.21 (16--27) \\
    Counselor utterance, chars     & 100.5 / 76 \\
    Student utterance, chars       & 59.3 / 57 \\
    Distinct session lengths       & 12 \\
    Empty / role-prefixed utterances & 0 / 0 \\
    \bottomrule
  \end{tabular}
  \caption{Scale of the dialogues. Utterance lengths are mean / median; ranges in parentheses.}
  \label{tab:scale}
\end{table}

\begin{table*}[t]
  \centering\small
  \begin{tabular}{@{}L{0.035\textwidth}L{0.2\textwidth}L{0.22\textwidth}L{0.18\textwidth}L{0.25\textwidth}@{}}
    \toprule
    & \textbf{Design commitment} & \textbf{Observable} & \textbf{Null} & \textbf{Result} \\
    \midrule
    C1 & Session length scales with the event's stress level & Pearson $r$ between stress level and turn count & Shuffle stress levels across sessions & $r=0.668$; null $0.000\pm0.027$; $p<0.001$ \\
    C2 & Student affect follows the event's stress level & Density of distress and absolutist markers in student utterances & Shuffle stress levels across sessions & $r=0.153$; null $-0.001\pm0.029$; $p<0.001$ \\
    C3 & Counselor follows the within-session phases & Counselor utterance length and question density per phase & --- & 39 / 80 / 139 median chars; 0.94 / 0.83 / 0.88 questions \\
    C4 & Dependency edges carry into the dialogue & Share of an earlier event's anchor terms recalled in a later session & Shuffle edge labels within gap strata; cross-student pairing & 3.53\% vs.\ 1.10\%; chance 0.69\%; positive in 8/8 strata; $p=0.001$ \\
    C5 & Later sessions retrieve earlier content & Sessions with cross-session reference; traceability of quotations & --- & 91.8\% of non-first sessions; 92.4\% of attributed quotations match the student's words; 45 of 109 earlier-session quotations match earlier wording \\
    C6 & The counselor never sees the core conflict & Substrings of \code{core\_conflict} in counselor turns & --- & 0 / 1{,}248 sessions \\
    \bottomrule
  \end{tabular}
  \caption{Dialogue conformance. Each row compares a property fixed in the input with a property measured on the generated dialogue. C4 is stratified by session gap because dependency edges concentrate at short gaps; the stratified null is centred at $+0.566$\,pp ($\pm0.090$), and the observed $+2.44$\,pp exceeds every resample.}
  \label{tab:conformance}
\end{table*}

\begin{figure*}[t]
  \centering
  \includegraphics[width=\textwidth]{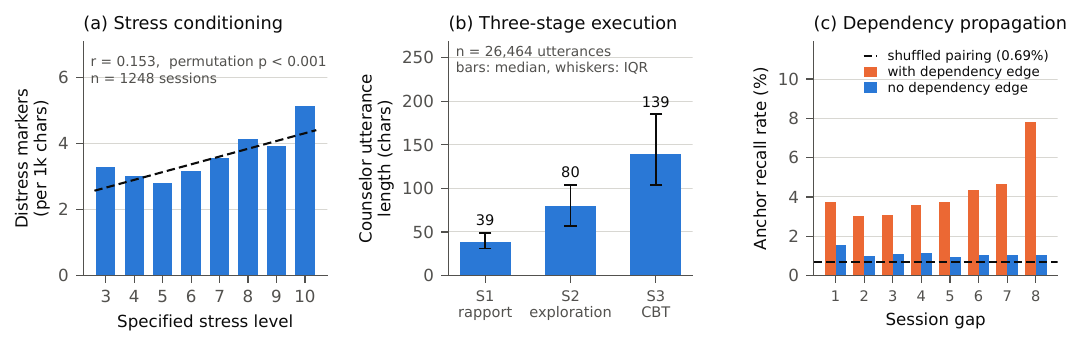}
  \caption{Conformance of the dialogues. (a) Distress and absolutist markers in student utterances by the event's stress level; the dashed line is a least-squares fit over all sessions. (b) Counselor utterance length by within-session phase; bars give the median, whiskers the interquartile range. (c) Recall of an earlier event's anchor terms in a later session, by dependency edge and session gap; the dashed line is the cross-student shuffled baseline.}
  \label{fig:conformance-full}
\end{figure*}

\begin{table}[t]
  \centering\footnotesize
  \begin{tabular}{@{}crrrr@{}}
    \toprule
    \textbf{Gap} & \textbf{Linked} & \textbf{$n$} & \textbf{Unlinked} & \textbf{$n$} \\
    \midrule
    1 & 3.75 & 671 & 1.53 & 477 \\
    2 & 3.04 & 411 & 0.99 & 637 \\
    3 & 3.09 & 295 & 1.12 & 653 \\
    4 & 3.57 & 147 & 1.12 & 701 \\
    5 & 3.77 & 102 & 0.96 & 646 \\
    6 & 4.36 &  50 & 1.04 & 598 \\
    7 & 4.67 &  17 & 1.05 & 531 \\
    8 & 7.83 &  22 & 1.04 & 426 \\
    \midrule
    All & 3.53 & 1{,}715 & 1.10 & 4{,}669 \\
    \bottomrule
  \end{tabular}
  \caption{Anchor recall (\%) by session gap for session pairs linked or unlinked by a dependency edge. Gaps beyond 8 have fewer than 15 linked pairs and are excluded from the strata and the totals; the margin of +2.44\,pp is computed from unrounded means.}
  \label{tab:gap}
\end{table}

\paragraph{Stress conditioning (C1, C2).}
C1 checks a mechanism the pipeline sets directly, as session length is computed from the stress level. C2 checks whether the student agent follows an instruction, as its prompt asks for emotional expression consistent with the stress level (Appendix~\ref{app:prompt-student}). Distress and absolutist markers rise from 3.29 per thousand characters at level~3 to 5.13 at level~10 (Figure~\ref{fig:conformance-full}a). The trend is not strictly monotone (the lowest band, 2.82, is at level~5) and the correlation is weak, so the student agent follows this instruction only loosely.

\paragraph{Within-session phases (C3).}
Counselor utterances lengthen across the three main phases, establishment and exploration, deep exploration, and CBT intervention (39, 80, and 139 median characters), while question density is highest in the first, matching the phase instructions: open questions first, longer restructuring responses later.

\paragraph{Dependency propagation (C4).}
For each event we take the character 3- to 4-grams of its content that occur in no other event of the same student, and measure what share of them reappear in each later session. Sessions linked to an earlier event by a dependency edge recall 3.53\% of its anchors, against 1.10\% for unlinked pairs and 0.69\% for a cross-student pairing. The difference is positive at every gap and generally widens with distance, from $+2.22$\,pp at a gap of one session to $+6.79$\,pp at eight (Table~\ref{tab:gap}). Both agents are given the antecedents of each session's event (Appendix~\ref{app:prompts}), so C4 measures how far the dependency edges carry into the dialogue; whether a model draws such connections without being given them is tested by the dynamic protocol (Section~\ref{sec:rq4}). The two widest strata rest on 17 and 22 linked pairs, so the right end of this trend is imprecise.

\paragraph{Cross-session reference and information asymmetry (C5, C6).}
Cross-session reference appears in 91.8\% of non-first sessions, with 1.47 counselor and 0.87 student reference turns per session. Coverage alone does not rule out a counselor attributing to the student something the student never said, so we trace quotations in a random sample of 25 students. Of 3{,}674 counselor quotations explicitly attributed to the student, 88.0\% match the student's wording in the same session, 4.4\% in an earlier session, and 0.7\% the profile or the week's event; 6.9\% match none of these. The 109 quotations that the counselor presents as coming from an earlier session are the cross-session case: 45 match earlier wording, 45 match wording from the current session, 1 matches the event description, and 18 match nothing. On reading all 18, none invents an episode absent from the trajectory. They are counselor self-quotations, paraphrases of earlier remarks in the same session, or reworded summaries of events in the graph. This bounds fabrication rather than proving its absence. The counselor prompt contains no field for the student's \code{core\_conflict} (Appendix~\ref{app:prompt-counselor}), and as a check on the text, no counselor turn contains a substring of it.

\paragraph{Risk-related content.}
Crisis intervention is outside the scope of CPCD, but some health events, such as episodes of disordered eating, approach it. We screened all events and sessions for suicide- and self-harm-related expressions with a keyword list. The screen finds four events, two of which describe passive death wishes, and three sessions from two students in which such expressions occur; in these sessions the counselor explores the expression but does not assess risk. We mark these sessions in the release so that they can be excluded.

\paragraph{Redundancy and stylistic concentration.}
No two sessions are near-duplicates at the surface level: over all 778{,}128 session pairs, the character 4-gram Jaccard similarity of the counselor side never reaches $0.10$ (maximum $0.098$), and the six most similar pairs all come from the same student, five of them adjacent sessions. Opening turns are distinct in 1{,}240 of 1{,}248 sessions. Per utterance, the counselor's similarity to its own earlier utterances with the same student is 1.16, 1.63, and 1.90 times its similarity to other students in the three phases, so its wording becomes more specific to the student as the session moves into intervention. Student utterances are less repetitive than counselor utterances throughout. The counselor nonetheless has recurring formulations: the most frequent appears in 8.5\% of the 26{,}464 counselor utterances. Because character $n$-gram counts split one long formulation across several overlapping windows, we report this deduplicated rate rather than raw $n$-gram rankings.

% ==================================================================
\section{Calibration of the Evaluation Pipeline}
\label{app:calib}
% ==================================================================

\begin{table}[t]
  \centering\footnotesize
  \begin{tabular}{@{}lccc@{}}
    \toprule
    \textbf{Metric} & \textbf{Reported} & \textbf{Ours} & \textbf{Diff.} \\
    \midrule
    Helpfulness & 4.99 & 4.98 & $-0.01$ \\
    Empathy     & 4.98 & 4.99 & $+0.01$ \\
    Guidance    & 4.98 & 4.98 & \phantom{$+$}0.00 \\
    Coherence   & 4.99 & 5.00 & $+0.01$ \\
    \midrule
    WAI task    & 4.11 & 4.04 & $-0.07$ \\
    WAI bond    & 4.69 & 4.98 & $+0.29$ \\
    WAI goal    & 4.44 & 4.05 & $-0.39$ \\
    WAI average & 4.41 & 4.36 & $-0.05$ \\
    \bottomrule
  \end{tabular}
  \caption{Scores reported for the MusPsy corpus in its original paper and scores obtained by our pipeline on the released corpus (GPT-4o, 100 sessions).}
  \label{tab:calib}
\end{table}

To check our implementation of the protocol, we scored the released MusPsy corpus and compared the result with the scores reported by \citet{wang-etal-2026-psychological} (Table~\ref{tab:calib}). The basic dimensions reproduce within 0.01 and the WAI average within 0.05. The WAI subscales do not reproduce: mass moves from goal to bond while the average is preserved. The MusPsy paper describes three items per subscale in the main text but lists four in its appendix, so the published subscales may follow a different item assignment from the one we use (goal: items 4, 6, 8, 11; task: 1, 2, 10, 12; bond: 3, 5, 7, 9). We therefore compare WAI subscales only within our own pipeline.

% ==================================================================
\section{CBT Adherence}
\label{app:ctrs}
% ==================================================================

The CTRS~\citep{young1980ctrs} is an observer-rated instrument of therapist competence in CBT. Its 11 items, each scored 0--6, cover general counseling skills such as understanding and collaboration as well as session structure such as agenda setting and homework; \citet{lee2024cactus} used it to assess synthetic CBT counseling. GPT-4o rates the same sessions as in Table~\ref{tab:quality}, using item descriptions we wrote from the scale definitions rather than the original manual, so our scores are not directly comparable with those of \citet{lee2024cactus}.

\begin{table}[t]
  \centering\footnotesize
  \begin{tabular}{@{}lc@{}}
    \toprule
    \textbf{Item group} & \textbf{Mean (0--6)} \\
    \midrule
    All 11 items                         & 4.90 \\
    Six general items$^{\dagger}$        & 5.60 \\
    \midrule
    Agenda                               & 4.00 \\
    Feedback                             & 3.50 \\
    Homework                             & 4.00 \\
    \bottomrule
  \end{tabular}
  \caption{Cognitive Therapy Rating Scale scores of CPCD sessions (GPT-4o, 100 sessions). $^{\dagger}$Understanding, interpersonal effectiveness, collaboration, guided discovery, focus on key cognitions, and strategy for change, the subset used by \citet{lee2024cactus}.}
  \label{tab:ctrs}
\end{table}

Table~\ref{tab:ctrs} reports the scores. The six general items used by \citet{lee2024cactus} lie close to the top of the scale, so they separate sessions little. We therefore report the three items tied to session structure separately: setting an agenda, eliciting feedback, and assigning homework. These are also the items that connect consecutive sessions, since the agenda opens by reviewing the previous assignment and the homework carries into the next session.

% ==================================================================
\section{Ablation Settings}
\label{app:ablation}
% ==================================================================

All ablated settings are generated for the same 100 students with the same agents, random seeds, and number of sessions per student as the full setting, so every session has a counterpart in each setting.

\paragraph{W/o dependencies.}
Each student keeps the same events in the same weeks, and all dependency edges are removed. The student agent therefore receives no antecedent events, and the counselor's session goal no longer names the earlier events the session builds on (Appendix~\ref{app:prompts}). Session goals and the pace of counseling are otherwise derived from the event of each session as in the full setting. Event descriptions rarely restate their antecedents: of an earlier event's character 4-grams, 0.47\% reappear in the description of a linked later event and 0.21\% in that of an unlinked one, so this setting removes nearly all explicit dependency information. As anchor terms occur in no other event of the same student, anchors recalled in this setting come from the dialogue or the memory, not from the current event.

\paragraph{W/o timeline.}
On top of w/o dependencies, the events are randomly reassigned to the weeks of the original graph, so a student meets the same events in a different order. Because each description was written after its antecedents, a reassigned event may refer to circumstances that have not yet occurred. Dependency recall is measured against the edges of the original graph; on average half of the linked pairs have their antecedent placed after the current event and are excluded.

\paragraph{W/o profile.}
The profile keeps gender, age, academic year, and major. The background, personality type, traits, coping style, and core conflict are removed for both agents. The event graph is unchanged. Because the events were generated from the full profile, their descriptions may still carry some of the removed information, so this setting likely underestimates the contribution of the profile.

% ==================================================================
\section{Training Details}
\label{app:training}
% ==================================================================

\begin{table}[t]
  \centering\footnotesize
  \begin{tabular}{@{}lr@{}}
    \toprule
    Base model           & Qwen2.5-7B-Instruct \\
    Fine-tuning          & LoRA \\
    LoRA rank / $\alpha$ & 32 / 16 \\
    Learning rate        & $2\times10^{-4}$ \\
    Epochs               & 2 \\
    Framework            & LlamaFactory \\
    Training seeds       & 3 or 1$^{\dagger}$ \\
    \bottomrule
  \end{tabular}
  \caption{Training configuration shared by all fine-tuned models. $^{\dagger}$Three seeds for CPCD-Chat, w/o dependencies, and w/o timeline; one for the others.}
  \label{tab:training}
\end{table}

All fine-tuned models share the configuration in Table~\ref{tab:training} and differ only in their training data. The external corpora are down-sampled to the number of counselor tokens in CPCD, so that every model sees a comparable amount of counselor speech. Single-session corpora contribute only response generation; multi-session corpora contribute the three tasks below.

\paragraph{Task construction.}
The three roles of memory, planning, and response have been used to train multi-session counselors on corpora such as MusPsy~\citep{wang-etal-2026-psychological}, where the goal of each session follows the therapy plan of a published case report and can largely be anticipated from earlier sessions. In CPCD, the focus of each session is a new event, which the counselor cannot know before the client reports it. We keep the three roles but build every input and target from information available at that point, using the same procedure for all multi-session corpora:
\begin{itemize}[leftmargin=*,itemsep=1pt,topsep=2pt]
  \item \emph{Memory update.} Input: the memory after session $t{-}1$ and the transcript of session $t$. Target: the record of session $t$, regenerated with the memory prompt (Appendix~\ref{app:prompt-memory}) from the transcript alone, without the event description used during generation.
  \item \emph{Planning.} Input: the memory after session $t$. Target: a plan written by the generator from that memory alone, stating the CBT stage reached, the homework to review, and topics left open. The plan does not contain the stage adjustment that the pipeline derives from the stress of the next event, so the counselor must read the pace from the client during the session.
  \item \emph{Response.} Input: the memory after session $t{-}1$, the plan for session $t$ built as above, and the dialogue so far. Target: the counselor's next utterance. During generation, the counselor's session goal is derived from the event graph (Appendix~\ref{app:prompt-counselor}), and in 41\% of sessions the counselor names the week's event before the student does, almost always when proposing the agenda. These turns (2.5\% of counselor turns) remain in the context but are excluded from the loss, so that no model is trained to state an event the client has not yet reported.
\end{itemize}
The shared module of the evaluation protocols writes memory and plans with the same procedure. MusPsy is translated into Chinese with DeepSeek-V4-Flash before the three tasks are built from its sessions, memories, and plans. We did not assess translation quality separately, and translation errors may put this baseline at a disadvantage.

% ==================================================================
\section{Dynamic Protocol: Metrics and End-to-End Results}
\label{app:dynamic}
% ==================================================================

\paragraph{Metric definitions.}
\emph{D1 dependency recall} measures whether the counselor connects the week's event to the events it depends on: the anchor recall of Section~\ref{sec:rq1}, computed on counselor utterances, for linked minus unlinked earlier events at the same gap. \emph{D2 stress responsiveness} measures whether the counselor adapts its pace to the client's state: the judge labels each counselor utterance as stabilization, exploration, cognitive restructuring, or assignment, and we report the mean Spearman correlation between the share of stabilization and the week's stress level. D1 and D2 are the pre-specified primary metrics. As secondary metrics, \emph{D3} reports how often the counselor follows up on the week's event once the client mentions it (uptake) and asks about the client's week when the client has not raised it within four turns (proactive inquiry); \emph{D4} reports the share of the counselor's quotations attributed to earlier sessions that cannot be found in the simulated history; \emph{D5} reports the change in the simulated client's affect within each session, measured by the Positive and Negative Affect Schedule (PANAS)~\citep{watson1988panas} before and after the session; and \emph{D6} applies the quality metrics of Section~\ref{sec:setup} to each session. Judge-based metrics require the judge to quote supporting evidence before giving a label.
D1 is computed on counselor utterances with the anchor terms of Section~\ref{sec:rq1}, and D4 counts as fabricated a quotation attributed to an earlier session for which no source can be found in the simulated history. Because D4 relies on string matching, it misses paraphrased fabrications and is a lower bound. D5 uses the PANAS item means (1--5) reported by the simulated client immediately before and after each session, and we group sessions by the tercile of the week's stress level.

\begin{table}[t]
  \centering\footnotesize
  \setlength{\tabcolsep}{3.5pt}
  \begin{tabular}{@{}lccccc@{}}
    \toprule
    \textbf{Model} & \textbf{P} & \textbf{R} & \textbf{D1} & \textbf{D2} & \textbf{D6} \\
    \midrule
    MusPsy       & \textbf{0.89} & 0.71 & +0.6 & 0.18 & 4.62 \\
    w/o dependencies & \textbf{0.89} & 0.71 & +0.7 & 0.31 & 4.65 \\
    w/o timeline & 0.88 & 0.70 & +0.5 & 0.30 & 4.64 \\
    w/o profile  & 0.87 & 0.69 & +1.4 & 0.28 & 4.61 \\
    \textbf{CPCD-Chat} & \textbf{0.89} & \textbf{0.72} & \textbf{+1.6} & \textbf{0.33} & \textbf{4.66} \\
    \bottomrule
  \end{tabular}
  \caption{Dynamic protocol in end-to-end mode, where each model updates its own memory and plans its own sessions. P/R: precision and recall of the facts in the model's memory against what the client said in earlier sessions; D1 in pp, D2 as Spearman $\rho$, D6 as the basic average.}
  \label{tab:e2e}
\end{table}

\paragraph{End-to-end mode.}
In the main comparison, memory and plans come from a shared module so that models differ only in their responses. Table~\ref{tab:e2e} lets each model trained on multi-session data maintain its own memory and plan its own sessions. The five models keep memories of similar accuracy, since memory records what the client said regardless of the timeline. Dependency recall, however, separates them more than in the shared-memory mode: CPCD-Chat reaches +1.6\,pp against +0.7\,pp for w/o dependencies, +0.5\,pp for w/o timeline, and +0.6\,pp for MusPsy; the gap to w/o dependencies is 0.9\,pp, compared with 0.5\,pp in Table~\ref{tab:cross}. Since plans contain no information about the coming event, the larger gap reflects what each model records in its own memory and which open topics its plans carry forward. Session quality drops slightly for all five models, reflecting errors that accumulate in self-maintained memory.

% ==================================================================
\section{Human Evaluation}
\label{app:human}
% ==================================================================

\paragraph{Raters.}
Two psychology experts, each holding a bachelor's degree and with at least two years of relevant work experience, rated all items independently. They were not involved in building the pipeline, were blind to the source of each session (corpus or model) and to all automatic scores, and received a one-hour briefing with five practice sessions not used in the analysis.

\paragraph{Dataset quality.}
We rated the 100 CPCD sessions scored by GPT-4o in Table~\ref{tab:quality} and 100 MusPsy sessions, using the Chinese translation of MusPsy described in Appendix~\ref{app:training} so that both corpora are rated in the same language, in a mixed random order. Raters used the same four basic dimensions and the same twelve WAI-SR items as the judge, each on a 1--5 scale, and we average the two raters. Table~\ref{tab:human-quality} compares expert and judge scores. Inter-rater agreement (Krippendorff's $\alpha$, interval) is 0.62 for the basic average and 0.67 for the WAI average.

\begin{table}[t]
  \centering\footnotesize
  \setlength{\tabcolsep}{3pt}
  \begin{tabular}{@{}lcccc@{}}
    \toprule
    & \multicolumn{2}{c}{\textbf{CPCD}} & \multicolumn{2}{c}{\textbf{MusPsy}} \\
    \cmidrule(lr){2-3}\cmidrule(l){4-5}
    \textbf{Metric} & \textbf{GPT-4o} & \textbf{Expert} & \textbf{GPT-4o} & \textbf{Expert} \\
    \midrule
    Basic average & 4.95 & 4.74 & 4.99 & 4.80 \\
    WAI goal      & 3.95 & 3.89 & 4.05 & 4.18 \\
    WAI task      & 3.90 & 3.86 & 4.04 & 4.12 \\
    WAI bond      & 4.80 & 4.70 & 4.98 & 4.63 \\
    WAI average   & 4.22 & 4.15 & 4.36 & 4.31 \\
    \midrule
    Spearman $\rho$, basic & \multicolumn{2}{c}{0.48} & \multicolumn{2}{c}{0.45} \\
    Spearman $\rho$, WAI   & \multicolumn{2}{c}{0.57} & \multicolumn{2}{c}{0.53} \\
    \bottomrule
  \end{tabular}
  \caption{Expert and GPT-4o ratings of the same 100 sessions per corpus. $\rho$: session-level correlation between the judge and the mean of the two experts.}
  \label{tab:human-quality}
\end{table}

Experts score both corpora lower than the judge, by 0.21 and 0.19 on the basic average, while preserving its ordering. The WAI subscales of the experts restore part of the goal advantage of MusPsy that our judge pipeline moves into bond (Appendix~\ref{app:calib}). On the basic average, the standard deviation across sessions is 0.38 for the experts and 0.09 for the judge.

\paragraph{Cross-session integration.}
From the dynamic protocol, we sampled 50 student--week pairs whose event has at least one antecedent, stratified over the five trajectory archetypes, and took the CPCD-Chat and w/o dependencies sessions for each (seed 1), presented in random order. For each session, raters marked (i) whether the counselor connected the week's event to an earlier event (none, correct, or incorrect, judged against the event graph shown to the raters), (ii) who first raised the connection (counselor or client), and (iii) how appropriate the connection was for the session (1--5, only when a connection was made); for each pair they then stated a preference or a tie. Disagreements on labels (i) and (ii) were resolved by discussion; a pair counts as a tie when the raters prefer different sessions. Agreement before discussion is $\alpha=$ 0.66 for (i), 0.72 for (ii), and 0.61 for (iii).

\begin{table}[t]
  \centering\footnotesize
  \setlength{\tabcolsep}{4pt}
  \begin{tabular}{@{}lcc@{}}
    \toprule
    & \textbf{CPCD-Chat} & \textbf{w/o dep.} \\
    \midrule
    Correct connection (\%)          & 58 & 34 \\
    Incorrect connection (\%)        & 4  & 6  \\
    No connection (\%)               & 38 & 60 \\
    Raised by counselor first (\%)   & 34 & 12 \\
    Appropriateness (1--5)           & 3.9 & 3.6 \\
    \midrule
    Preferred pairs (of 50)          & 31 & 13 \\
    \bottomrule
  \end{tabular}
  \caption{Expert labels for 50 paired sessions from the dynamic protocol. Percentages are over the 50 sessions of each model; appropriateness averages the sessions with a connection. Six pairs are ties.}
  \label{tab:human-integration}
\end{table}

Table~\ref{tab:human-integration} reports the labels. The gap is largest for connections the counselor raises before the client mentions the earlier event (34\% vs.\ 12\%), so the advantage is not explained by repeating what the simulated client has just said. Appropriateness is higher for CPCD-Chat (Wilcoxon rank-sum $p=$ 0.04). Across all 100 sessions, D1 anchor recall is higher for sessions labeled as correctly connected (point-biserial $r=$ 0.41).

% ==================================================================
\section{Generation Prompts}
\label{app:prompts}
% ==================================================================

All prompts were issued in Chinese; we give English translations. Fields in braces are filled at run time. The format instructions produced by the output parser and the repair note appended when a generated event graph fails validation are omitted.

\subsection{Student Profile}
\label{app:prompt-profile}

\begin{promptbox}{Prompt 1: Student profile generation}
[System]
You are a psychology research assistant who creates realistic, specific, and mutually distinct psychological profiles of university students.
Requirements: reflect the real circumstances of contemporary Chinese university students; avoid stereotypes; make the psychological distress specific and deep; write in Chinese.
[JSON schema of the profile, generated by the output parser; omitted]

[User]
Generate a detailed psychological profile of a university student.

**The following attributes are fixed by the study design and must be used exactly as given:**
- Name: {name}
- Gender: {gender}
- Year of study: {year}
- Major: {major}
- MBTI: {mbti}
- Primary stress domain: {domain}

**Description of this stress domain**: {domain_desc}

**Parts for you to write**:
1. age: an age consistent with {year}.
2. personality.traits: a personality description consistent with {mbti}; do not copy the official MBTI descriptions.
3. personality.coping_mechanism: a clear stress-coping mechanism (e.g., avoidant, aggressive, suppressive, overcompensating, ...).
4. background: family and social background. It must include factors that may pose psychological risk (left-behind childhood, financial hardship, harsh parenting, son preference, single-parent family, etc.). Give the specific region, family structure, and details of growing up; do not write a generic template.
5. core_conflict: summarize in one sentence the contradiction the student is most torn by. It must be closely related to [{domain}] and traceable to the background.

**Diversity requirement**: This is one entry in a dataset of over a hundred profiles. Avoid the most common settings (e.g., high-frequency combinations such as "from a remote mountain village, relying on student loans, parents favor sons"), and give this student's situation recognizably individual features.

Output the JSON directly, strictly following the structure of the JSON example above, without Markdown code-block markers.
\end{promptbox}

\noindent\code{\{name\}}, \code{\{gender\}}, \code{\{year\}}, \code{\{major\}}, \code{\{mbti\}} and \code{\{domain\}} are assigned by the sampling design of Appendix~\ref{app:profile} and overwritten in the returned JSON. \code{\{domain\_desc\}} is the description of the domain in Table~\ref{tab:domains}.

\begin{table}[t]
  \centering\footnotesize
  \begin{tabularx}{\columnwidth}{@{}L{1.7cm}X@{}}
    \toprule
    \textbf{Domain} & \textbf{Typical stressors / \emph{potential psychological correlates}} \\
    \midrule
    Academic pressure & Failing courses or a falling GPA; failing the postgraduate entrance exam or missing a recommendation; a rejected paper; being let down by teammates in group work; delayed graduation; being unable to follow lectures. \emph{Anxiety, reduced self-efficacy, learned helplessness, shame} \\
    Interpersonal relationships & Dormitory conflicts (schedules, hygiene); being excluded or isolated; a breakup; unrequited love; social-anxiety episodes; conflict with a teacher. \emph{Loneliness, interpersonal sensitivity, depressed mood, anger, sense of abandonment} \\
    Career development & Being dismissed from an internship; no offers in autumn recruitment; failing the civil-service exam; no interest in one's major; aimlessness; peer ``involution'' pressure. \emph{Confusion, existential anxiety, social-comparison pressure, emptiness} \\
    Family and finances & Parental divorce; serious illness of a relative; insufficient living expenses; family debt; value conflicts with parents; trauma from a left-behind childhood. \emph{Lack of security, guilt, low self-worth, sense of deprivation} \\
    Physical and mental health & Severe insomnia; appearance anxiety; weight or eating disorders; internet addiction; sudden illness; accidental injury. \emph{Somatization, self-loathing, impulse-control problems} \\
    \bottomrule
  \end{tabularx}
  \caption{Stress domain descriptions supplied as \code{\{domain\_desc\}}.}
  \label{tab:domains}
\end{table}

\subsection{Temporal Stress Event Graph}
\label{app:prompt-event}

\begin{promptbox}{Prompt 2: Temporal stress event graph generation}
[System]
You are a campus-life simulator that generates structured data on students' stress events. Output valid JSON only.

[User]
Based on the [Student Profile], generate a temporal stress event graph covering one semester for this student.

[Student Profile]
{persona}

[Structural specification for this generation -- must be strictly satisfied]
1. Number of events: exactly {n_events}.
2. Time span: the first event is in week {week_start} and the last in week {week_end}; the remaining events fall between these two weeks. All events are sorted by week in ascending order.
3. Domain quota (only the following domains may be used, each with the specified number of events):
{quota_lines}
4. Stress trajectory shape: {shape_desc}
5. Stress-intensity constraint (paired with the trajectory shape above; must be satisfied): {stress_rule_text} A relief event can be a small positive turn, support from others, completion of one stage of a task, or simply a relatively calm week -- do not let stress escalate monotonically from beginning to end.
6. Event IDs are E1, E2, ..., E{n_events}, numbered in chronological order.
7. caused_by may only reference the IDs of events in **earlier weeks**. Except for the first event, most events should have a causal antecedent, forming traceable chains.

[Content requirements]
- The events must reflect the [core conflict] in the profile and be consistent with the student's family background, personality, and coping mechanism.
- Events must be concrete: name the course, the relationships, and the details of the scene; do not write vague descriptions such as "academic pressure increased".
- The primary domain {primary_domain} is the main thread of this trajectory. Events in secondary domains should be entangled with the main thread rather than form an independent list.
- psychological_impact must describe a specific psychological reaction, not just the word "anxiety".

[Output format]
Return a JSON object with a single key, events, whose value is the list of events:

{
  "events": [
    {
      "id": "E1",
      "week": {week_start},
      "domain": "{primary_domain}",
      "event_content": "a concrete event description with identifiable details",
      "psychological_impact": "a description of the specific psychological reaction",
      "stress_level": 5,
      "caused_by": []
    }
  ]
}

Do not output Markdown code-block markers; return the JSON directly.
\end{promptbox}

\noindent\code{\{persona\}} is the full profile. \code{\{n\_events\}}, \code{\{week\_start\}}, \code{\{week\_end\}}, \code{\{primary\_domain\}} and \code{\{quota\_lines\}} follow the specification of Appendix~\ref{app:event}. \code{\{shape\_desc\}} is the description of the sampled archetype in Table~\ref{tab:shape-desc}, and \code{\{stress\_rule\_text\}} states its constraint from Table~\ref{tab:shapes} in words. The field \code{caused\_by} holds the dependency edges of Section~\ref{sec:method} and is released as \code{depends\_on}.

\begin{table}[t]
  \centering\footnotesize
  \begin{tabularx}{\columnwidth}{@{}lX@{}}
    \toprule
    \textbf{Archetype} & \textbf{\code{\{shape\_desc\}}} \\
    \midrule
    Escalating    & Stress accumulates gradually overall and peaks in the later part of the semester, with one or two brief reliefs along the way. \\
    Spike--relief & An acute peak occurs in mid-semester, followed by a clear decline with support or self-adjustment; the final part stays at a moderate level. \\
    Oscillating   & Stress rises and falls repeatedly; peaks and troughs alternate with no single highest point. \\
    Late onset    & The early part of the semester is relatively calm (mostly low-to-moderate stress); the triggering event appears in the middle-to-late part and stress then rises quickly. \\
    Early crisis  & A strong shock occurs at the start of the semester, followed by slow recovery in mid-semester; stress rises again at the end due to new real-world pressures. \\
    \bottomrule
  \end{tabularx}
  \caption{Trajectory archetype descriptions supplied as \code{\{shape\_desc\}}.}
  \label{tab:shape-desc}
\end{table}

\subsection{Student Agent}
\label{app:prompt-student}

\begin{promptbox}{Prompt 3: Student agent}
### Role: Simulated Client (Student Agent)
You are now the university student **{name}**.

### Persona
- Basic information: {demographics}
- Personality and coping style: {personality}
- Background: {background}
- Core conflict (you may not be able to put it into words yourself, but it governs your reactions): {core_conflict}

### Current Context
- Time: week {current_week} (counseling session {session_id} of this semester)
- Key event this week: {event_content}
- Event domain: {event_domain}
- Current psychological stress level (1-10): {stress_level}
- Related prior experiences: {prior_summary}
- Current psychological reaction: {psychological_impact}

### Your memory of previous counseling sessions
{memory_summary}

### Instructions
1. State-driven: your emotions and words must be strictly constrained by the "current event", "psychological reaction", and "stress level". When the stress level is low, do not appear more distressed than you are; when it is high, do not appear overly calm.
2. Gradual disclosure: do not reveal everything at the start. Show realistic defense mechanisms (denial, rationalization, changing the subject, downplaying). Open up gradually, only after the counselor has built trust.
3. Memory recall: when the counselor mentions, or you yourself recall, content from past sessions, respond coherently based on the memory above. You may bring it up yourself, e.g., "I tried that method you suggested last time." Do not invent past counseling episodes that are not in your memory.
4. Do not state the core conflict directly: you would never say "my core conflict is ..."; it shows only indirectly through your hesitation, avoidance, and ambivalence.
5. Dialogue only: output only what you say. Strictly no parenthetical action descriptions (e.g., "(sighs)"), emoji, inner monologue, or speaker-name prefixes.
6. Brief and realistic: this is spoken conversation, usually 1-4 sentences; do not write it as a written paragraph.

### Current dialogue
Counselor: "{last_counselor_response}"
Generate your reply:
\end{promptbox}

\noindent The prompt is filled again at every student turn. The profile fields come from Prompt~1 and the event fields from the event of the current session. \code{\{prior\_summary\}} lists the events this event depends on as \code{[Week w: event\_content]}, or reads ``None (this is a relatively independent new experience)''. \code{\{memory\_summary\}} renders the memory records of earlier sessions (Prompt~5), or reads ``This is the first session; no history yet.'' \code{\{last\_counselor\_response\}} is the counselor's latest utterance.

\subsection{Counselor Agent}
\label{app:prompt-counselor}

\begin{promptbox}{Prompt 4: Counselor agent}
### Role: Expert Psychological Counselor (Counselor Agent)
You are conducting session {session_id} with the student {student_name}
(week {week} of this semester).

### Context
- Basic profile: {basic_info}
- Memory of previous sessions: {memory_summary}
- Dialogue history of this session:
{dialogue_history}

### Session Goal
{session_goal}

### Last Assignment
{last_assignment}

### Cross-session progression: Stage {cbt_stage} -- {cbt_stage_name}
Aims of this stage: {cbt_stage_aims}
{cbt_technique_block}

### Current within-session phase: {current_phase}

### Action instruction for this phase
{phase_instruction}

### Absolute dialogue principles
1. Ask only one question at a time: never pose several questions in one reply.
2. Socratic guidance: do not tell the client directly "how you should think"; use questions to guide them to their own conclusions (except during crisis intervention).
3. Keep it brief and conversational: natural and colloquial, no lengthy theoretical exposition.
4. Follow the phase instruction strictly: do not perform tasks outside the current phase.
5. Do not repeat what you have said before: move forward from a different angle or with different wording.
6. Output only what you say, with no speaker-name prefix and no parenthetical descriptions.

Reply as the counselor:
\end{promptbox}

\noindent The prompt is filled again at every counselor turn. \code{\{memory\_summary\}} is the same as in Prompt~3, the stage fields are given in Appendix~\ref{app:prompt-cbt}, and the phase fields in Appendix~\ref{app:prompt-phase}. The remaining fields are filled as follows. The counselor prompt contains no core conflict, background, personality, or stress level.

\begin{promptbox}{Prompt 4 (continued): Field templates}
{basic_info}
Name: {name}, Gender: {gender}, Age: {age}, Year: {year}, Major: {major}

{session_goal}
Center the session on this week's event ({event_content, first 60 characters}...) and work through the {psychological_impact, first 40 characters} it triggered; task for this stage: {cbt_stage_aims}.
[If the event has antecedents in the event graph, append:]
Also build on the earlier "{antecedent event_content, first 24 characters; up to two antecedents}", helping the client see the connection between them.

{last_assignment}
The assignments field of the previous session's memory record (P5), or "(No assignment was given last time)".

{cbt_technique_block}
[Stages 2-6:]
CBT techniques available at this stage (pick one that fits the moment; do not list them and do not name the technique):
- {technique name} | {technique description}
[Stage 1:]
This stage focuses on building rapport and assessment; no specific cognitive technique is used yet.
\end{promptbox}

\subsection{Memory Consolidation}
\label{app:prompt-memory}

\begin{promptbox}{Prompt 5: Memory consolidation}
### Role: Memory Consolidation Module
Generate a structured summary of this counseling session, to be stored in the long-term memory bank and retrieved in later sessions.

### Input
- Student: {student_name}
- Core event of this session (week {week}): {event_content}
- Dialogue transcript:
{dialogue_transcript}

### Requirements
- Be specific: name the specific people, scenes, and wordings discussed in this session; do not write empty phrases such as "discussed academic pressure".
- counselor_intervention must name the techniques actually used (e.g., Socratic questioning, cognitive restructuring, behavioral experiment, emotion labeling) and the topics they addressed.
- student_progress must describe verifiable changes or signs of resistance.
- session_goal and assignments must faithfully reflect content that actually appeared in the dialogue; leave them empty if absent, and do not fill them in.
- Keep summary_text within 200 characters; it is read before the next session begins.

### Output
Output strictly in JSON, with no explanatory text or code-block markers:
{
  "session_id": "{session_id}",
  "week": {week},
  "emotional_state": "the student's main emotional state",
  "key_topics": "key topics discussed in this session (specific)",
  "counselor_intervention": "main interventions used by the counselor",
  "student_progress": "the student's change or resistance",
  "session_goal": "the focus this session actually addressed (the agreed focus if one was explicitly agreed at the start; otherwise the focus actually discussed)",
  "assignments": "the exact wording of the specific exercise assigned to the client in this session; an empty string if none was assigned",
  "summary_text": "a concise summary for reference in the next session (within 200 characters)"
}
\end{promptbox}

\noindent The prompt is called once after each session. The fields \code{emotional\_state}, \code{key\_topics}, \code{counselor\_intervention}, \code{student\_progress} and \code{summary\_text} form \code{\{memory\_summary\}} in later sessions; \code{assignments} fills \code{\{last\_assignment\}} of the next session, and \code{summary\_text} fills \code{\{last\_summary\}} of the opening prompt.

\subsection{Within-Session Phases}
\label{app:prompt-phase}

After the opening line (turn 0), the counselor speaks $T$ turns, where $T$ is $16+\mathrm{round}(0.9\,s)+u$, where $s$ is the stress level of the session's event and $u$ is drawn uniformly from $\{-2,\dots,2\}$, clipped to $[14,30]$. Table~\ref{tab:phases} assigns each counselor turn a phase.

\begin{table}[t]
  \centering\footnotesize
  \begin{tabularx}{\columnwidth}{@{}lX@{}}
    \toprule
    \textbf{Counselor turn} & \textbf{Phase} \\
    \midrule
    0 & Opening line (fixed greeting in session 1) \\
    1 & Agenda: Review \\
    2 & Agenda: Setting \\
    3--8 & Establishment \& Exploration \\
    9--12 & Deep Exploration \\
    13 to $T-1$ & CBT Intervention \\
    $T-2$, $T-1$ & scheduled phase + Wrap-up add-on \\
    $T$ & Closing \\
    \bottomrule
  \end{tabularx}
  \caption{Phase schedule within a session. In Prompt~4, the phase name fills \code{\{current\_phase\}} and the instruction of the same name below fills \code{\{phase\_instruction\}}.}
  \label{tab:phases}
\end{table}

\begin{promptbox}{Opening line and first-session greetings}
[Opening prompt -- sessions 2 onward]
(System) You are a professional Chinese psychological counselor.
(User)
You are a psychological counselor about to begin session {session_id} with the student {name}; {gap} week(s) have passed since the last session.

Summary of the last session:
{last_summary}

Write only one opening line (no more than 35 characters):
- colloquial, natural and brief, like the greeting of a real person walking into the room;
- you may briefly mention what was discussed last time or the small exercise that was assigned, but do not start probing for details right away;
- do not use stock phrases such as "How was your week?" or "How have you been feeling lately?";
- output only the line itself, with no quotation marks, no speaker-name prefix and no explanation.

[Session-1 greetings -- one sampled at random]
- Hello, {name}, welcome. I'm the counselor here. To begin, I'd like to hear what made you decide to come in.
- {name}, hello, please have a seat. We have fifty minutes, and you can start wherever you like.
- Hello, {name}. Since this is our first meeting, I'd like to hear it in your own words -- what has been happening lately that made you want to talk to someone?
- {name}, welcome. No need to organize your thoughts first; just say whatever comes to mind.
\end{promptbox}

\noindent\code{\{last\_summary\}} is the \code{summary\_text} of the previous session, and \code{\{gap\}} is the number of weeks since that session.

\begin{promptbox}{Phase instructions}
[Agenda: Review]
Current task: review the exercise from last time. Do only this; do not mention what to discuss today.
- If the "Last Assignment" above is not "(No assignment was given last time)", ask concretely how it went -- name the exercise itself rather than using a stock question such as "How have things been?".
- If no exercise was assigned (including the first session), bring the client's attention to the present in one sentence; do not invent an exercise.
One or two sentences, natural and colloquial.

[Agenda: Setting]
Current task: set today's agenda. Do only this.
1. In one sentence the client can understand, propose the focus suggested for today (based on the "Session Goal" above); do not copy the goal verbatim.
2. Explicitly ask for agreement and wait for the client's response, e.g., "Shall we start from here today?"
Note: you already asked about last time's exercise in the previous turn -- **do not ask again**; and do not start exploring in this turn -- proposing a focus and obtaining agreement is enough.
One or two sentences.

[Establishment & Exploration]
Current task: only listen and clarify the problem.
1. Your reply must be brief (no more than 2 sentences).
2. Use minimal encouragers (e.g., "Mm", "I see", "That sounds hard").
3. Use open-ended questions to guide the student to describe in detail what happened.
4. Strictly no analysis or advice at this stage.

[Deep Exploration]
Current task: understand the student's emotions in depth.
1. Empathize with and reflect the emotions the student expresses.
2. Ask about specific situational details and clarify vague information.
3. Try to connect the current event with the memory of previous sessions (if the memory contains related content, cite it specifically rather than vaguely referring to "before").

[CBT Intervention]
Current task: give a professional, compassionate and helpful response. Build on the structure of a cognitive-behavioral therapy response, paying particular attention to "identifying key thoughts or beliefs", and connect the parts naturally:
1. Validation and empathy: show understanding of the client's feelings or problem and create a sense of safety.
2. Identify key thoughts or beliefs: find possible cognitive distortions or core beliefs in the client's narrative.
3. Challenge or reflect: use open questions to encourage the client to re-examine this thought.
4. Offer strategies or insights: give actionable strategies to help the client handle the current situation.
5. Encourage and look ahead: encourage the client to practice, and note that this is only a beginning and further support may be needed.
Note: these five points are a structure completed gradually over the session; do not cover all five in a single reply.

[Wrap-up add-on]
### Wrap-up instruction (this session is nearing its end)
After completing the task above, begin to wrap up: do not open any new direction of exploration; bring the conversation back to what has already been discussed today, and prepare for ending.

[Closing]
Current task: this is the last reply of this session; you must end the session in this reply.
1. In one or two sentences, summarize the core content discussed today and tie it back to the direction agreed at the start.
2. Assign a small, concrete exercise that can be done next week, and **explain why** -- which thought or difficulty discussed today it relates to, and what the client might notice after doing it. Assigning without a rationale does not count as complete.
3. Solicit feedback: ask how today's conversation felt -- what helped, what did not feel right, or what was not covered. Ask sincerely, not as a formality.
4. End with a clear goodbye, so the client knows today's session is over and that you will meet again.
Never raise any new exploratory question, and do not leave any topic open.
\end{promptbox}

\subsection{Cross-Session CBT Stages}
\label{app:prompt-cbt}

Table~\ref{tab:stages} lists the six stages of the counseling course and the techniques available in each; stage names and aims fill \code{\{cbt\_stage\_name\}} and \code{\{cbt\_stage\_aims\}} in Prompt~4, and the techniques, with the descriptions of Table~\ref{tab:techniques}, fill \code{\{cbt\_technique\_block\}}. The stage of each session is planned from the event graph before the session is generated. The stress level is used only for planning and is not shown to the counselor agent.

\begin{table}[t]
  \centering\footnotesize
  \begin{tabularx}{\columnwidth}{@{}L{2.2cm}X@{}}
    \toprule
    \textbf{Stage} & \textbf{Aims / \emph{techniques}} \\
    \midrule
    1. Building Trust and Assessment & Build the relationship; assess the current situation; provide psychoeducation. \\
    2. Identifying Negative Cognitions & Track automatic thoughts; link emotions with thoughts; identify cognitive distortions. \emph{Evidence-Based Questioning; Pie Chart Technique} \\
    3. Challenging False Beliefs & Test the validity of thoughts; generate alternative perspectives; carry out reality testing. \emph{Decatastrophizing; Reality Testing; Pros and Cons Analysis; Efficiency Evaluation} \\
    4. Restructuring Cognitive Patterns & Restructure negative thoughts; develop adaptive responses; verify new perspectives. \emph{Alternative Perspective; Continuum Technique; Changing Rules to Wishes} \\
    5. Behavioral Skill Building & Behavioral activation; graded exposure; problem-solving training. \emph{Behavior Experiment; Problem-Solving Skills Training; Systematic Exposure} \\
    6. Consolidation and Termination & Summarize gains; prevent relapse; end the counseling formally. \emph{Efficiency Evaluation} \\
    \bottomrule
  \end{tabularx}
  \caption{Counseling stages across sessions. The stages follow the course of CBT described by \citet{beck2011cbt}, and the techniques are taken from the catalogue of \citet{lee2024cactus}; the assignment of techniques to stages is ours.}
  \label{tab:stages}
\end{table}

\begin{table}[t]
  \centering\footnotesize
  \begin{tabularx}{\columnwidth}{@{}lX@{}}
    \toprule
    \textbf{Technique} & \textbf{Description} \\
    \midrule
    Efficiency Evaluation & Examine what a thought or behavior actually helps with and what it costs. \\
    Pie Chart Technique & Divide the responsibility for an event among all parties in proportion, to break ``it is all my fault''. \\
    Alternative Perspective & Generate several other equally plausible explanations of the same situation. \\
    Decatastrophizing & Follow ``what is the worst that could happen, then what, and how could you cope'' step by step. \\
    Pros and Cons Analysis & List the benefits and costs of holding a thought or keeping a behavior. \\
    Evidence-Based Questioning & Ask what evidence supports the thought and what evidence contradicts it. \\
    Reality Testing & Design a small step that tests the expectation in reality. \\
    Continuum Technique & Place an all-or-nothing judgment on a 0--100 continuum. \\
    Changing Rules to Wishes & Rewrite ``I must \dots'' as ``I wish \dots''. \\
    Behavior Experiment & Design a small trial to test whether a belief holds. \\
    Problem-Solving Skills Training & Define the problem, generate options, evaluate them, choose one and act on it. \\
    Systematic Exposure & Approach avoided situations gradually along a hierarchy from easy to hard. \\
    \bottomrule
  \end{tabularx}
  \caption{Technique descriptions supplied to the counselor agent.}
  \label{tab:techniques}
\end{table}

\begin{promptbox}{Stage planning}
Input: session index i (1-based), number of sessions N,
stress level s of the session's event,
highest stage reached so far r (initially 1).

planned = min(6, max(1, ceil(6 * i / N)))
if i == N:    stage = 6;            mode = advance
elif s >= 8:  stage = max(1, planned-1); mode = hold
elif s <= 4:  stage = min(6, planned+1); mode = accelerate
else:         stage = planned;      mode = advance
floor = r - 1 if mode == hold else r
stage = min(6, max(stage, floor, 1))
r = max(r, stage)
\end{promptbox}

\noindent The last session always consolidates; an acute peak ($s\ge8$) holds or steps back one stage, and a calm period ($s\le4$) moves ahead. Progress is never undone except for this single step back.

\end{document}